\documentclass[10pt,twocolumn,letterpaper]{article}

\usepackage[pagenumbers]{cvpr} 

\newcommand{\Input}{\State \textbf{Input:} }
\newcommand{\HyperParam}{\State \textbf{HyperParam:} }

\usepackage{algorithm}
\usepackage{algpseudocode}
\usepackage{amsmath} 
\usepackage{multirow}

\definecolor{cvprblue}{rgb}{0.21,0.49,0.74}
\usepackage[pagebackref,breaklinks,colorlinks,allcolors=cvprblue]{hyperref}

\def\paperID{9977} 
\def\confName{CVPR}
\def\confYear{2026}

\title{Learning to Expand Images for Efficient Visual Autoregressive Modeling}

\author{Ruiqing Yang$^1$, Kaixin Zhang$^2$, Zheng Zhang$^3$, Shan You$^4$, Tao Huang$^{5}$$\thanks{Corresponding author.}$\\
$^1$University of Electronic Science and Technology of China \\
$^2$School of Computer Science and Engineering, Central South University \\
$^3$Xidian University\quad
$^4$SenseTime Research\quad
$^5$Shanghai Jiao Tong University\\
\tt\small yrq@std.uestc.edu.cn, kaixinzhang@csu.edu.cn,  \\
\tt\small zheng.zhang@stu.xidian.edu.cn, youshan@sensetime.com, t.huang@sjtu.edu.cn}

\begin{document}
\maketitle
\begin{abstract}
Autoregressive models have recently shown great promise in visual generation by leveraging discrete token sequences akin to language modeling. However, existing approaches often suffer from inefficiency, either due to token-by-token decoding or the complexity of multi-scale representations. In this work, we introduce Expanding Autoregressive Representation (EAR), a novel generation paradigm that emulates the human visual system’s center-outward perception pattern. EAR unfolds image tokens in a spiral order from the center and progressively expands outward, preserving spatial continuity and enabling efficient parallel decoding. To further enhance flexibility and speed, we propose a length-adaptive decoding strategy that dynamically adjusts the number of tokens predicted at each step. This biologically inspired design not only reduces computational cost but also improves generation quality by aligning the generation order with perceptual relevance.Extensive experiments on ImageNet demonstrate that EAR achieves state-of-the-art trade-offs between fidelity and efficiency on single-scale autoregressive models, setting a new direction for scalable and cognitively aligned autoregressive image generation.Code is available at \url{https://github.com/RuiqingYoung/EAR}.
\end{abstract}    
\section{Introduction}
\label{sec:intro}

Inspired by the remarkable success of large language models (LLMs)~\cite{wan2023efficient} in natural language processing, their autoregressive next-token prediction strategy has been extended to the field of image generation, driving substantial progress in autoregressive visual modeling.

\begin{figure}[t]
\centering
\includegraphics[width=0.9\columnwidth]{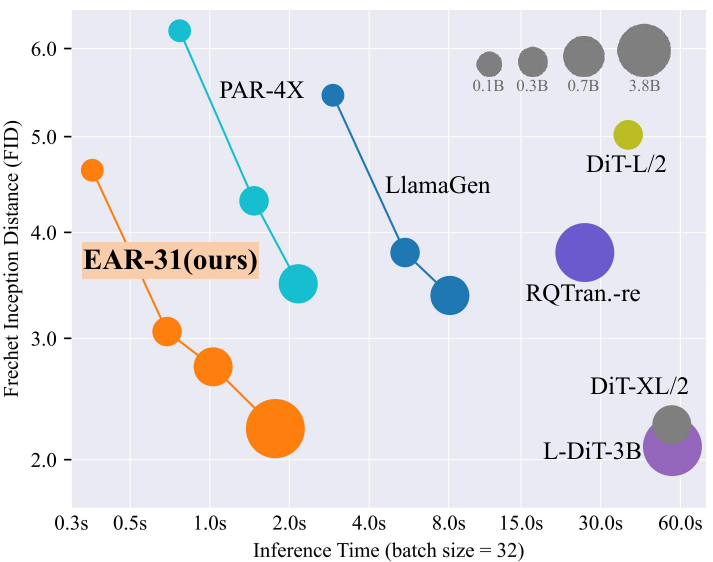} 
\caption{Scaling behavior of different generation methods on ImageNet $256\times256$ generation benchmark.}
\label{Scaling behavior}
\end{figure}

\begin{figure*}[t]
\centering
\includegraphics[width=0.8\linewidth]{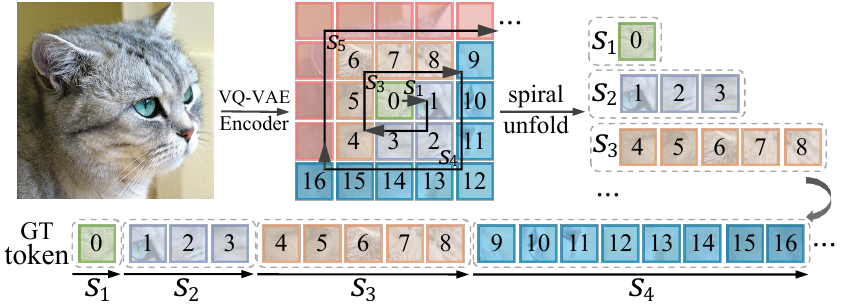} 
\caption{The proposed spiral unfolding and parallel generation strategy. The generation starts from the center of an image, and expands outwards spirally. Each ${S}$ stands for a generation step. }
\label{fig:spiral_unfolding}
\end{figure*}

Recent autoregressive image generation methods follow the design of large language models (LLMs), where an image is first encoded into a sequence of discrete visual tokens using a tokenizer such as VQ-VAE~\cite{van2017neural}. These tokens are then autoregressively generated one by one. However, since generating a single image typically requires a large number of tokens (e.g., $16 \times 16$), the next-token prediction paradigm leads to significantly more inference steps compared to diffusion models~\cite{gu2021vectorquantizeddiffusion}, resulting in much slower generation speed. Although several recent works such as PAR~\cite{wang2025parallelizedautoregressivevisualgeneration} and MAR~\cite{li2025autoregressiveimagegenerationrandomized} have attempted to accelerate autoregressive generation by patch-wise or randomized generation strategies, they often suffer from increased model complexity or limited flexibility. More recently, VAR achieves a significant reduction in inference steps through a next-scale prediction mechanism, attaining competitive performance in both accuracy and speed compared to diffusion models. However, it introduces substantial computational overhead during inference and requires training multi-scale VQ-VAE tokenizers. This raises an important question: \textit{is there an image modeling paradigm that not only obeys the nature of image understanding to obtain promising generative capabilities, but also retains the simplicity and efficiency of autoregressive generation?}

In this paper, inspired by insights from cognitive neuroscience, we propose a novel image generation framework called Expanding Autoregressive Modeling (EAR). To emulate the human visual system’s center-outward scanning mechanism, we introduce a novel spiral unfolding strategy that preserves the spatial continuity among image tokens. As illustrated in Figure~\ref{fig:spiral_unfolding}, unlike the conventional raster-scan-based token arrangements, in our method, each token remains adjacent to its spatial neighbors after unfolding, enabling a center-to-outward generation process that more naturally aligns with the inherent spatial structure of images.

Meanwhile, to better allocate the generation resources, we design a progressive length-varying decoding strategy: in the challenging early stages, the model predicts fewer tokens to ensure a reliable start of the central region; as generation proceeds outward, the number of tokens per step increases accordingly to boost the efficiency.
As a result, our EAR obtains several major advantages over previous methods:
\begin{itemize}
    \item \textbf{Preserved spatial continuity:} Unlike multi‑scale prediction methods (e.g., VAR) or patch‑chunk generation strategies (e.g., PAR), spiral ordering avoids disrupting adjacency among visual tokens, thereby improving image fidelity and reducing local artifacts. 
    \item \textbf{Efficient parallel expansion:} Initiating generation from the center and expanding outward enables controlled parallel prediction of multiple tokens per step, resulting in significantly faster inference while preserving generation quality.
    \item \textbf{Biologically inspired perceptual alignment:} The center‑to‑outward generation pattern aligns with human visual processing mechanisms, including central fixation bias and center–surround attention profiles~\cite{Linka2025}~\cite{Boehler2009}), promoting cognitive plausibility in the generated images.
\end{itemize}

We implement a series of EAR model variants on $256\times256$ ImageNet generation. The results demonstrate that, our EAR achieves the optimal balance between image quality and generation speed, as summarized in Figure~\ref{Scaling behavior}. For example, our EAR31-XL model achieved an FID of 2.7 and runs 8 times faster than LlamaGen with the same number of parameters.

\section{Related Work}
\label{sec:RelatedWork}

\subsection{Autoregressive Models in Image Generation}
Autoregressive models have been widely adopted in image generation by drawing parallels between language token modeling and visual data synthesis. Early studies~\cite{reed2016generating, salimans2017pixelcnn++, lee2022autoregressive, zheng2022movq} propose the next-pixel prediction paradigm, where each pixel is treated as a token. These methods often rely on flattening 2D images into 1D token sequences via raster scan order~\cite{van2016conditional, chen2020generative, chen2018pixelsnail, lee2022autoregressive}, and employe CNNs or Transformers to predict next pixel based on all previous pixels. In order to improve the quality and efficiency of image generation, VQVAE and VQGAN~\cite{van2017neural, esser2021taming} introduce the tokenizer that compresses image patch into discrete visual tokens.  Building on this foundation, recent works~\cite{wang2024omnitokenizer, zhu2024scaling, mattar2024wavelets} make substantial progress, achieving image generation quality on par with diffusion-based models~\cite{rombach2022high, podell2023sdxl, peebles2023scalable}. For example, LLamaGen~\cite{sun2024autoregressive} redesigns the image tokenizer to support multiple downsampling ratios, resulting in higher image generation quality.

\subsection{Acceleration of Autoregressive Models}
In natural language processing~\cite{brown2020language, devlin2019bert} and computer vision~\cite{dosovitskiy2020image, touvron2021training, liu2021swin}, the transformer~\cite{vaswani2017attention} has become one of the most popular backbone networks, owing to its scalability and strong performance. However, this performance often comes with long inference latency, which poses a challenge for the efficient deployment. To alleviate this issue, many studies~\cite{santacroce2023matters, ma2023llm, zhu2021vision, yu2023x, lou2024token}propose acceleration techniques to improve inference speed of transformer. Motivated by these efforts, some researchers~\cite{li2024autoregressive, he2024zipar, shen2025numerical} begin exploring acceleration strategies specifically for transformer-based autoregressive models. For instance, Anagnostidis et al.\cite{anagnostidis2023dynamic} introduce a context-aware pruning method that dynamically removes uninformative tokens during generation. DD~\cite{liu2024distilled} proposed a method based on knowledge distillation, which significantly reduced the inference steps of the autoregressive model. 
Despite acceleration efforts, most methods sacrifice quality, depend on task-specific heuristics, or lack generality across token structures and decoding schemes. Many also target language models and struggle with high-res image generation, where spatial coherence is vital. To address this, we propose EAR, a spatially aware framework with a biologically inspired center-outward generation and flexible next-any-tokens prediction. This enables parallel multi-token generation, speeding inference while preserving image quality and structure for efficient, high-fidelity synthesis.
\section{Learning to Expand Images}
\label{sec:method}

\begin{figure}[t]
    \centering
    \includegraphics[width=0.6\linewidth]{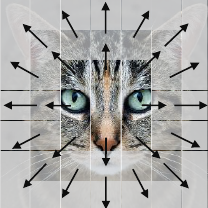}
    \caption{Illustration of human visual perception. The observations fixate initially at the center of an image, then gradually expand outwards.}
    \label{fig:human}
\end{figure}

\begin{figure*}[t]
\centering
\includegraphics[width=1\textwidth]{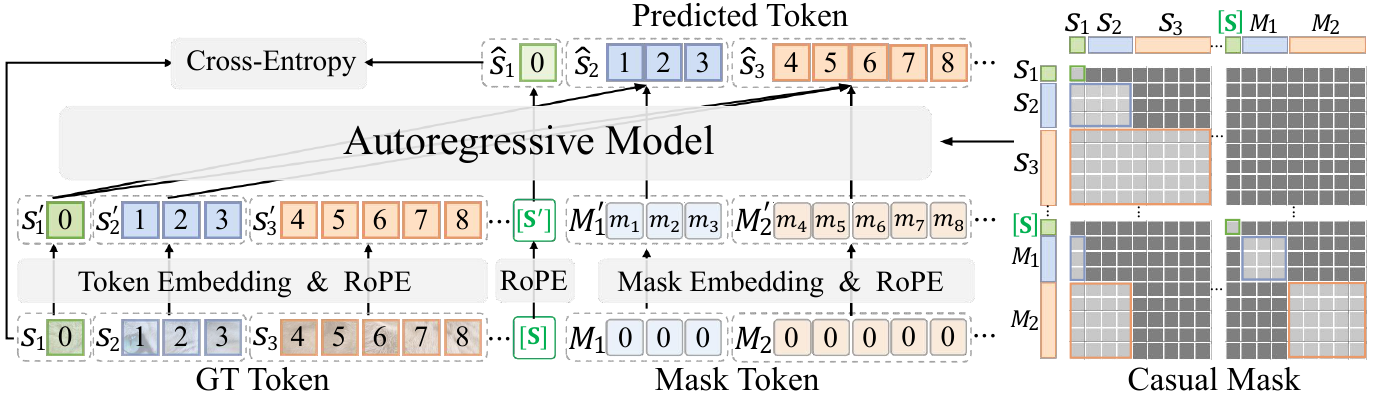} 
\caption{Training EAR transformer on tokens. Note that [S] denotes the start token derived from the class embedding. In practice, the Mask Token also attends to [S] through the attention mechanism to capture class information, although this detail is omitted in the figure for clarity.}
\label{model}
\end{figure*}

\subsection{Preliminary on Visual Autoregressive Modeling}

Current visual autoregressive (AR) generation methods can be categorized based on the prediction type into next-token prediction and next-scale prediction.

\subsubsection{Next-token Prediction}

Autoregressive models decompose the joint distribution of a token sequence \(x = (x_1, x_2, \ldots, x_T)\) into a product of conditional probabilities:

\begin{equation}
    p(x_1, x_2, \ldots, x_T) = \prod_{t=1}^{T} p(x_t \mid x_1, x_2, \ldots, x_{t-1})
\end{equation}

This formulation assumes each token depends only on its previous tokens. It has been widely successful in language modeling and has recently been applied to visual domains by tokenizing images into sequences (e.g., using raster-scan order). However, sequential token generation limits speed, particularly for high-resolution images.

\subsubsection{Next-scale Prediction}

To improve generation efficiency, the next-scale prediction paradigm introduces a hierarchical token generation process. Rather than modeling individual tokens, it organizes token maps across multiple spatial resolutions (scales) and generates them in a coarse-to-fine manner:

\begin{equation}
    p(r_1, r_2, \ldots, r_S) = \prod_{s=1}^{S} p(r_s \mid r_1, \ldots, r_{s-1})
\end{equation}

Here, \(r_s\) denotes the token map at scale \(s\), which can represent a downsampled version of the image. This approach allows the model to capture global structure first and then refine local details. In hierarchical models like VAR~\cite{li2024autoregressive}, generating a high-resolution image (e.g., 256×256) typically involves decoding multiple scales (e.g., 10 levels). Multi-scale token maps are concatenated and jointly input to the model, which increases computation due to longer token sequences and multi-scale attention. Additionally, all intermediate token maps need to be stored during generation, further increasing GPU memory consumption.

\subsection{Human Cognition Inspired Visual Modeling}

In this paper, we try to figure out two major questions: (1) What is the more natural and precise image understanding manner? (2) What parallelization of token generation is better in both quality and speed?

To answer the questions, we refer to human visual perception research in cognitive neuroscience (see an illustration in Figure~\ref{fig:human} for easier understanding), where studies consistently show that observers tend to fixate initially at the center of an image—a phenomenon known as the central fixation bias~\cite{Linka2025}. Meanwhile, magnetoencephalographic (MEG) experiments reveal that attention in early visual cortex exhibits a center‑surround profile, where focal enhancement at the center is surrounded by a narrow inhibitory zone, reflecting recurrent processing and feature binding during early feedforward stages~\cite{Boehler2009}. This indicates that human perception expands progressively from the center to outside, with each expansion step depending on its surrounding regions.

These motivates us to build a more efficient image modeling manner by starting from the center, and gradually expanding outward. Meanwhile, a straightforward parallelization can be built inside this expansion process.

Next, we will officially introduce the two cores of our method: \textbf{spiral token unfolding} and \textbf{next-any-tokens parallelization}.

\subsubsection{Spiral Token Unfolding}
Inspired by how the human visual system prioritizes fine-grained perception in the fovea and gradually expands attention to the periphery, we propose a novel token unfolding strategy called \textbf{spiral unfolding}, which preserves spatial continuity and enables center-outward image generation. Specifically, we start from the central token at $(\frac{n}{2}, \frac{n}{2})$ and traverse the 2D token grid in a clockwise spiral (right → down → left → up), producing a \textit{spiral index map} that defines the token order, as shown in Figure~\ref{fig:spiral_unfolding}.

\subsection{Next-Any-Tokens Parallelization}

\begin{figure}[h]
\centering
\includegraphics[width=1\columnwidth]{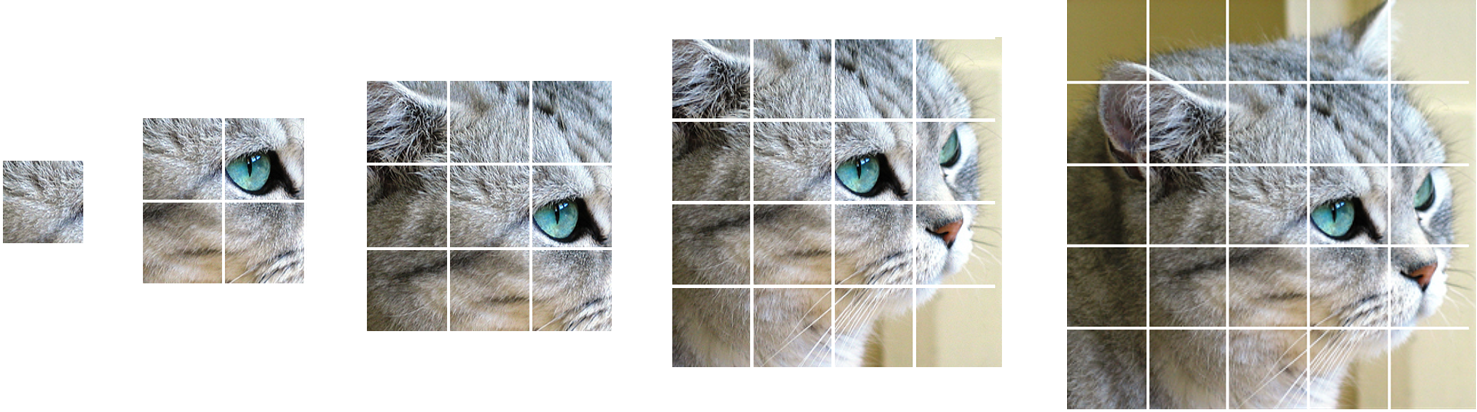} 
\caption{EAR generation process.}
\label{EAR Generation Process}
\end{figure}


After applying spiral unfolding to the image tokens, we obtain a token sequence that expands outward from the image center. To enable the model to efficiently generate a complete image following the expansion pattern shown in Figure~\ref{EAR Generation Process}, we design the Expandable Autoregressive Modeling (EAR) framework, and introduce a novel prediction paradigm: \textbf{next-any-tokens prediction}.

During the training, as illustrated in Figure~\ref{model}, we introduce a learnable embedding parameter similar to the mask token in MaskGit. This mask token is duplicated to match the spatial size of the ground truth (GT) tokens and then concatenated with the start token (obtained from the class embedding) and the GT tokens, as illustrated in Figure~\ref{model}. Both GT tokens and mask tokens are encoded with the same rotary positional encoding (RoPE). The resulting sequence is then fed into the autoregressive model along with our designed EAR causal mask, which ensures that the mask tokens are autoregressively transformed into predicted tokens. The predicted tokens represent a probability distribution over the codebook entries in VQ-VAE for each token position. Finally, we compute the loss between the predicted tokens and the GT tokens to train our EAR model. Therefore, the next-any-tokens prediction in our EAR framework can be formulated as:
\begin{equation}
    p(s_1, s_2, \ldots, s_T) = \prod_{t=1}^{T} p(m_t \mid s_1, s_2, \ldots, s_{t-1})
    \label{eq:next-any-tokens prediction}
\end{equation}
where $m_t$ denotes the mask tokens at the $t$-th step, and $s_1, s_2, \ldots, s_{t-1}$ are the ground truth tokens revealed in the previous steps.

\subsubsection{Mask Design}


To accommodate our next-any-tokens prediction paradigm, we design a customized causal mask. Specifically, for the mask tokens within the same generation step, we allow them to attend to all the ground truth (GT) tokens from previous steps, as well as to each other within the same step. This design prevents information leakage from future tokens while ensuring contextual consistency within the current step.

However, if no attention mask is applied to the GT tokens, the multi-layer structure of the Transformer may cause information from future GT tokens to be indirectly propagated to earlier-step GT tokens through residual connections and self-attention in deeper layers. To address this, we also apply attention masks to the GT tokens, restricting each GT token to only attend to GT tokens from the current and previous steps. An illustration of our EAR-specific causal mask design is shown in Figure~\ref{model} (right part).

\subsubsection{KV-Cache Design for Inference Acceleration}

\begin{figure}[t]
\centering
\includegraphics[width=1\columnwidth]{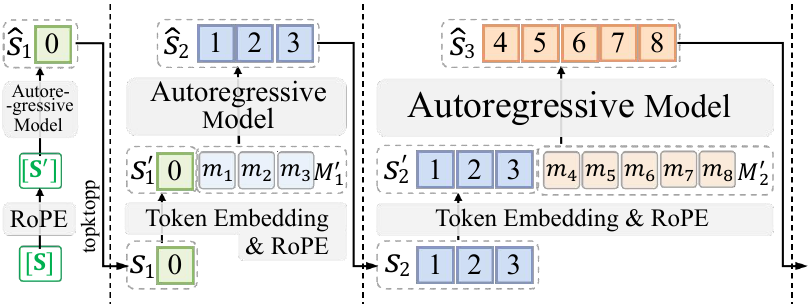} 
\caption{Inference process of EAR.}
\label{inference}
\end{figure}

To further accelerate our inference speed, we design a KV cache mechanism compatible with the next-any-tokens prediction paradigm. As shown in Figure~\ref{inference}, our autoregressive generation process begins with a start token $[S]$ derived from the class embedding. After passing through the trained model, it generates the first token $0$, and the corresponding key-value (KV) pairs of $[S]$ are stored in the KV cache.

In the second step, three mask tokens $m_1$, $m_2$, and $m_3$ are appended to the generated GT token $0$ and input into the model together. These mask tokens are transformed into GT tokens $1$, $2$, and $3$ respectively, and the KV pairs of tokens $0$, $m_1$, $m_2$, and $m_3$ are stored in the KV cache.

In the third step, five new mask tokens $m_4$, $m_5$, $m_6$, $m_7$, and $m_8$ are appended to the previously generated GT tokens $1$, $2$, and $3$, and passed through the model. These mask tokens are transformed into GT tokens $4$ through $8$. Importantly, the KV pairs of the newly generated GT tokens $1$, $2$, and $3$ overwrite the cached KV pairs of the previous mask tokens $m_1$, $m_2$, and $m_3$, while the KV pairs of $m_4$ through $m_8$ are appended to the KV cache.

Compared to traditional autoregressive models such as LlamaGen, our method maintains the same sequence length during inference and introduces no additional computational cost.

By iterating this process, we effectively avoid redundant computation of previously generated tokens' key-value pairs, thereby accelerating the image generation during inference.

\begin{algorithm}[ht]
\caption{Next-any-tokens Prediction Train}
\begin{algorithmic}[1]
\Input{latent image $gt$, class token $sos$}
\HyperParam{steps $K$, mask token $mt$, mask $M$}
\State $f = ARmodel(*),gt = \text{spiral}(gt)$
\State $MT = mt.\text{expand}(\text{len}(gt))$
\State $L_0 = \text{concat}(sos, gt, sos, MT)$
\For{$k = 1, \cdots, K$}
    \State $L_k = f(L_{k-1}) \gets M$
\EndFor
\State $logits \gets$ extract $MT$ positions from $L_K$
\State $loss = \text{CrossEntropy}(logits, gt)$\\
\Return{$loss$}
\end{algorithmic}
\end{algorithm}

\begin{algorithm}[ht]
\caption{Next-any-tokens Prediction Inference}
\begin{algorithmic}[1]
\Input{class token $sos$}
\HyperParam{steps $K$, tokens per step $(l_k)_{k=1}^K$, mask token $mt$, KV cache $C$}
\State $f = ARmodel(concat(*,*)) \to (*,*)$
\State $L_0 = sos, L=[\ ]$
\For{$k = 1, \cdots, K$}
    \State $MT_k = mt.\text{expand}(l_k)$
    \State $\_, L_k = f(L_{k-1},MT_k)\ and\ (L_{k-1},MT_k)\to C$
    \State $L = \text{queue\_push}(L,L_k)$
\EndFor
\State $L = \text{unspiral}(L)$\\
\Return{generated latent image $L$}
\end{algorithmic}
\end{algorithm}
\section{Experiments}

\begin{table*}[ht]
\centering
\footnotesize
\begin{tabular}{l l c c c c c c c c}
\toprule
\textbf{Type} & \textbf{Model} & \textbf{FID} $\downarrow$ & \textbf{IS} $\uparrow$ & \textbf{Pre} $\uparrow$ & \textbf{Rec} $\uparrow$ & \#Para & \#Step & \textbf{Time} $\downarrow$ & GFLOPs\\
\midrule
\multirow{6}{*}{Diff.}
& ADM~\cite{dhariwal2021diffusionmodelsbeatgans} & 10.94 & 101.0 & 0.69 & 0.63 & 554M & 250 & 265.75 & --\\
& CDM~\cite{JMLR:v23:21-0635} & 4.88 & 158.7 & -- & -- & -- & 8100 & -- & --\\
& LDM-4-G~\cite{rombach2022high} & 3.60 & 247.7 & -- & -- & 400M & 250 & -- & --\\
& DiT-L/2~\cite{peebles2023scalable} & 5.02 & 167.2 & 0.75 & 0.57 & 458M & 250 & 38.12 & --\\
& DiT-XL/2~\cite{peebles2023scalable} & 2.27 & 278.2 & 0.78 & 0.62 & 675M & 250 & 55.71 & --\\
& L-DiT-3B~\textit{(upgraded from DiT~\cite{peebles2023scalable})} & 2.10 & 304.4 & 0.82 & 0.66 & 3.0B & 250 & $>$55.71 & --\\
& L-DiT-7B~\textit{(upgraded from DiT~\cite{peebles2023scalable})} & 1.86 & 316.2 & 0.83 & 0.67 & 7.0B & 250 & $>$55.71 & --\\
\midrule
\multirow{3}{*}{VAR}
& VAR-d16~\cite{tian2024visualautoregressivemodelingscalable} & 3.30 & 274.4 & 0.84 & 0.51 & 310M & 10 & 0.49 & 214.74\\
& VAR-d20~\cite{tian2024visualautoregressivemodelingscalable} & 2.57 & 302.6 & 0.83 & 0.56 & 600M & 10 & 0.50 & 268.42\\
& VAR-d30~\cite{tian2024visualautoregressivemodelingscalable} & 1.92 & 323.1 & 0.82 & 0.63 & 2.0B & 10 & 1.24 & 402.64\\
\midrule
\multirow{3}{*}{AR}
& LlamaGen-B~\cite{sun2024autoregressive} & 5.46 & 193.6 & 0.84 & 0.46 & 111M & 256 & 2.92 & 25.97\\
& LlamaGen-L~\cite{sun2024autoregressive} & 3.80 & 248.3 & 0.83 & 0.52 & 343M & 256 & 5.47 & 93.42\\
& LlamaGen-XL~\cite{sun2024autoregressive} & 3.39 & 227.1 & 0.81 & 0.54 & 775M & 256 & 8.08 & 220.46\\
\midrule
\multirow{4}{*}{AR}
& NAR-B~\cite{he2025neighboringautoregressivemodelingefficient} & 4.65 & 212.3 & 0.83 & 0.47 & 130M & 31 & 0.50 & --\\
& NAR-M~\cite{he2025neighboringautoregressivemodelingefficient} & 3.27 & 257.5 & 0.82 & 0.47 & 290M & 31 & 0.71 & --\\
& NAR-L~\cite{he2025neighboringautoregressivemodelingefficient} & 3.06 & 263.9 & 0.81 & 0.53 & 372M & 31 & 0.83 & --\\
& NAR-XL~\cite{he2025neighboringautoregressivemodelingefficient} & 2.70 & 277.5 & 0.81 & 0.58 & 816M & 31 & 1.17 & --\\
\midrule
\multirow{3}{*}{AR}
& PAR-B-4X-2.19rFid$^*$~\cite{wang2025parallelizedautoregressivevisualgeneration} & 6.21 & 204.4 & 0.86 & 0.46 & 111M & 67 & 0.77 & 25.98\\
& PAR-L-4X-2.19rFid$^*$~\cite{wang2025parallelizedautoregressivevisualgeneration} & 4.32 & 189.4 & 0.87 & 0.43 & 343M & 67 & 1.47 & 93.44\\
& PAR-XL-4X-2.19rFid$^*$~\cite{wang2025parallelizedautoregressivevisualgeneration} & 3.50 & 234.4 & 0.84 & 0.49 & 775M & 67 & 2.16 & 220.50\\
\midrule
\multirow{3}{*}{AR (Ours)}
& EAR-B & 4.64 & 218.7 & 0.82 & 0.48 & 98M & 31 & 0.36 & 26.00\\
& EAR-L & 3.06 & 261.4 & 0.83 & 0.54 & 326M & 31 & 0.69 & 93.48\\
& EAR-XL & 2.75 & 275.1 & 0.83 & 0.56 & 754M & 31 & 1.03 & 220.58\\
\midrule
\multirow{3}{*}{AR (Ours)}
& EAR-B (adaLN) & 4.14 & 237.7 & 0.83 & 0.50 & 140M & 31 & 0.44 & 28.69\\
& EAR-L (adaLN) & 2.76 & 266.4 & 0.83 & 0.57 & 477M & 31 & 0.85 & 103.10\\
& EAR-XL (adaLN) & 2.54 & 262.7 & 0.83 & 0.57 & 
1.1B & 31 & 1.37 & 243.14\\

\bottomrule
\end{tabular}
\caption{Quantitative comparison of various generative models on ImageNet 256$\times$256. $*$: results implemented by NAR ~\cite{he2025neighboringautoregressivemodelingefficient} using a $16\times16$ tokenizer. To mitigate the difference between tokenizers, FLOPs is calculated only for transformer blocks for generation of one image.}
\label{Overall-comparison}
\end{table*}
\subsection{Experimental Setup}

To validate the effectiveness and scalability of the proposed EAR, we adopt a decoder-only Transformer architecture modified from the LlamaGen~\cite{sun2024autoregressive} implementation, and follow VAR~\cite{tian2024visualautoregressivemodelingscalable} designs by integrating AdaLN-based conditioning into the transformer blocks. All ablation experiments, however, are performed without AdaLN to ensure a clean comparison of architectural factors. By introducing spiral unfolding and the next-any-tokens prediction mechanism, EAR is capable of generating images from the center outward in significantly fewer steps.

\textbf{Class-conditional image generation.}
We evaluate EAR on the widely used ImageNet 256$\times$256 dataset. Images are tokenized using the pre-trained image tokenizers proposed by LlamaGen~\cite{sun2024autoregressive} and XQGAN~\cite{li2024xqganopensourceimagetokenization}, both with a downsampling factor of 16. We experiment with three generation steps: 10, 16, and 31. Specifically, the 16-step and 31-step settings correspond to per-step token counts of 
$[1, 3, 5, 7, \ldots, 31]$ (where the number of tokens at step $k$ is $2k-1$) and 
$[1, 1, 2, 2, \ldots, 15, 15, 16]$ (where the number of tokens at step $k$ is $ (k+1)//2 $), respectively. 
These arrays correspond to $(l_k)_{k=1}^K$ in the pseudocode.
Two configurations of mask tokens are used: one where a unified mask token is learned via a trainable embedding, and another where class tokens derived from class embeddings are used. All models are trained for 300 epochs with an initial learning rate of $2\times10^{-4}$ using a step-wise learning rate scheduler. Inception Score (IS) and Frechet Inception Distance (FID) are used as evaluation metrics, computed by sampling 50,000 images using the official TensorFlow evaluation toolkit provided by ADM~\cite{dhariwal2021diffusionmodelsbeatgans}.

\begin{figure*}[h]
\centering
\includegraphics[width=0.95\textwidth]{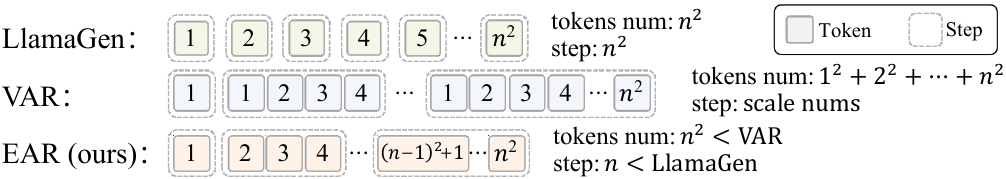} 
\caption{Comparisons of numbers of tokens at multiple generation steps among different methods.}
\label{token numbers and steps}
\end{figure*}

\begin{figure*}[h]
\centering
\includegraphics[width=1.0\textwidth]{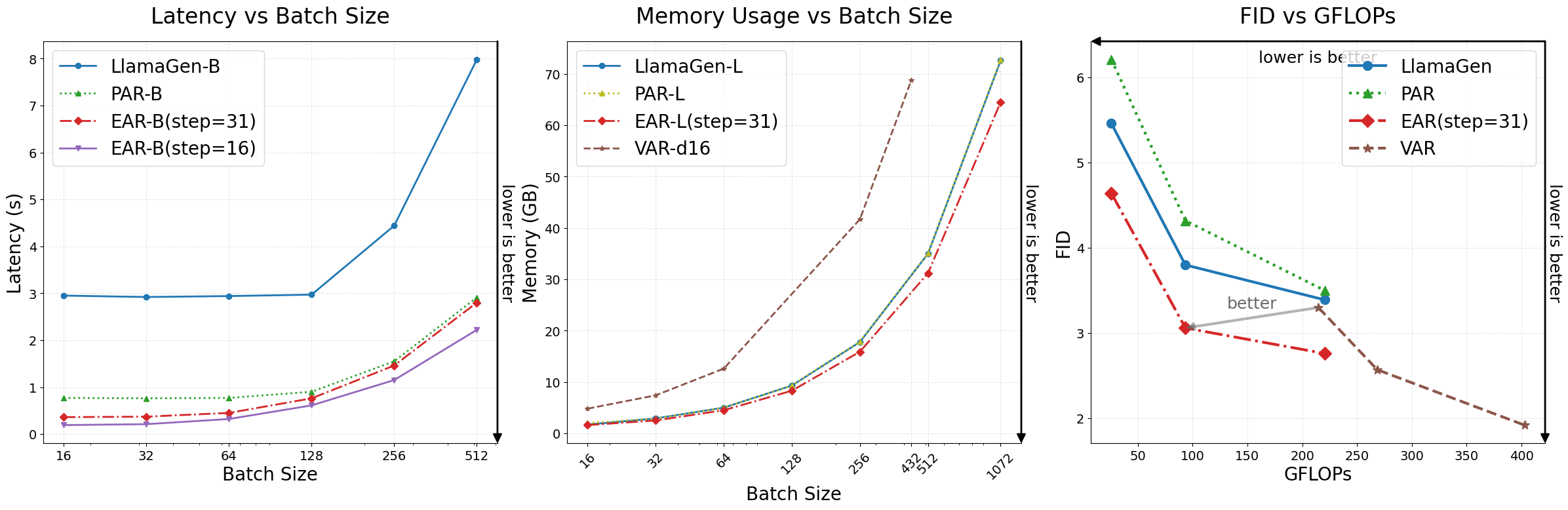} 
\caption{Efficiency comparisons between vanilla AR, PAR, VAR and the proposed EAR (without adaLN).}
\label{efficiency comparisons}
\end{figure*}

\subsection{Main Results}

We compare our proposed EAR models with recent state-of-the-art autoregressive image generation methods, including LlamaGen~\cite{sun2024autoregressive}, PAR~\cite{wang2025parallelizedautoregressivevisualgeneration}, NAR~\cite{he2025neighboringautoregressivemodelingefficient}, and scalable VAR models~\cite{tian2024visualautoregressivemodelingscalable}, as summarized in Table~\ref{Overall-comparison}.

Our EAR framework achieves a compelling balance between generation quality and computational efficiency. EAR-XL attains a strong FID of 2.54 with only 31 autoregressive steps, requiring just 1.37 seconds per 32-image batch and 220 GFLOPs. This represents a significant improvement over prior autoregressive approaches. For instance, LlamaGen-XL~\cite{sun2024autoregressive} achieves an FID of 3.39 but requires 256 steps and 8.08 seconds per batch, highlighting EAR's superior step efficiency and faster inference (83.0\% speedup) despite comparable parameter counts. 
Overall, our EAR models deliver better FID and IS scores at comparable parameter scales and inference speeds relative to the NAR~\cite{he2025neighboringautoregressivemodelingefficient} counterparts. For instance, EAR-B achieves a lower FID (4.14 vs. 4.65) than NAR-B while using a similar number of parameters (140M vs. 130M) and attaining a higher IS (237.7 vs. 212.3).
Notably, when compared with VAR~\cite{tian2024visualautoregressivemodelingscalable} models under a similar parameter scale, our EAR-L achieves a lower FID (2.76 vs. 3.30) and requires significantly fewer GFLOPs (103.10G vs. 214.74G).

These results demonstrate EAR's ability to deliver competitive or superior image quality while drastically reducing computational overhead, establishing a new efficiency benchmark for autoregressive visual generation. The synergy between spiral token unfolding and parallel next-any-tokens prediction enables high-fidelity synthesis without sacrificing scalability, addressing a core limitation of traditional autoregressive approaches.


\subsection{Comparison of Model Inference}

As shown in Figure~\ref{token numbers and steps}, suppose the token size of a single image is $n \times n$.
For LlamaGen, it needs to predict $n \times n$ tokens, and due to the next-token prediction paradigm, the number of inference steps required to generate an image is also $n \times n$.

For VAR~\cite{tian2024visualautoregressivemodelingscalable}, which follows the next-scale prediction paradigm, it needs to predict the sum of tokens across all scales. The number of inference steps corresponds to the number of scales used in the model, which is 10 as reported in the paper.

In contrast, our EAR model also only needs to predict $n \times n$ tokens, which is significantly fewer than the total number of multi-scale tokens used in VAR~\cite{tian2024visualautoregressivemodelingscalable}. 
Leveraging our proposed next-any-tokens prediction, we can generate an image in an arbitrary number of steps. 
To balance generation quality and speed, we empirically set the number of steps to $n$, which is significantly fewer than the $n \times n$ steps required by LlamaGen~\cite{sun2024autoregressive}. 
As illustrated in Figure~\ref{efficiency comparisons}, our inference time is significantly lower than that of LlamaGen~\cite{sun2024autoregressive}, and as the batch size increases, our memory usage remains far below that of VAR~\cite{tian2024visualautoregressivemodelingscalable}. Meanwhile, EAR achieves an excellent trade-off between FID and GFLOPs, demonstrating a well-balanced combination of generation quality and computational efficiency.

\subsection{Ablation Study}
\subsubsection{Effect of Step Numbers}

To investigate the relationship between decoding steps and generation quality, we conduct experiments using LlamaGen's tokenizer with 10 and 16 steps. As shown in Table~\ref{Effect-of-Step-Numbers1}, using 16 steps significantly improves image quality compared to 10 steps. Table~\ref{Effect-of-Step-Numbers2} also indicates that 31 steps achieve better results than 16 steps.

\begin{table}[h]
\footnotesize
\centering
\begin{tabular}{lccccc}
\toprule
Model & Params & FID & IS & Steps & Time (s) \\
\midrule
EAR-B  & 111M & 7.13 & 217.9 & 10 & 0.12 \\
EAR-B  & 111M & 6.39 & 231.8 & 16 & 0.21 \\
EAR-L  & 343M & 4.69 & 272.1 & 16 & 0.51 \\
EAR-XL & 775M & 3.90 & 285.0 & 16 & 0.88 \\
\bottomrule
\end{tabular}
\caption{Effect of different step numbers (with LlamGen's VQVAE~\cite{sun2024autoregressive}).}
\label{Effect-of-Step-Numbers1}
\end{table}

\begin{table}[h]
\footnotesize
\centering
\begin{tabular}{lccccc}
\toprule
Model & Params & FID & IS & Steps & Time (s) \\
\midrule
EAR-B  & 98M & 5.49 & 225.2 & 16 & 0.21 \\
EAR-L  & 326M & 3.55 & 256.44 & 16 & 0.51 \\
EAR-XL & 754M & 2.92 & 266.03 & 16 & 0.88 \\
EAR-B  & 98M & 4.64 & 218.7 & 31 & 0.36 \\
EAR-L  & 326M & 3.06 & 261.4 & 31 & 0.69 \\
EAR-XL & 754M & 2.75 & 275.1 & 31 & 1.03 \\
\bottomrule
\end{tabular}
\caption{Effect of different step numbers (with XQGAN's VQVAE~\cite{li2024xqganopensourceimagetokenization}).}
\label{Effect-of-Step-Numbers2}
\end{table}

\subsubsection{Effect of Mask Token Selection}

In the experimental stage, we explored two types of mask tokens: one derived from class embeddings as the category-specific start token, and the other being a unified, learnable mask embedding shared across all categories. As shown in the Table~\ref{mask_token_effect}, the results show that, under the same conditions, using the unified mask token achieves better performance than using the class embedding-derived mask tokens.

\begin{table}[h]
\footnotesize
\centering
\begin{tabular}{lcccc}
\toprule
Model & Params & FID & IS & Steps \\
\midrule
\multicolumn{5}{c}{\textit{Using Class Mask Token}} \\
\midrule
EAR-B & 111M & 6.78 & 209.3 & 16 \\
EAR-L & 343M & 4.88 & 238.8  & 16 \\
EAR-XL & 775M & 4.03 & 238.4  & 16 \\
\midrule
\multicolumn{5}{c}{\textit{Using Unified Mask Token}} \\
\midrule
EAR-B & 111M & 6.39 & 231.8  & 16 \\
EAR-L & 343M & 4.69 & 272.1  & 16 \\
EAR-XL & 775M & 3.90 & 285.0  & 16 \\
\bottomrule
\end{tabular}
\caption{Comparison of class-specific and unified mask tokens. Results indicate that unified mask tokens yield better performance across all model sizes.}
\label{mask_token_effect}
\end{table}

\subsection{Image Extension Task}
Thanks to our spiral token unfolding strategy, our EAR model can flexibly handle a variety of image extension tasks under different input conditions. In this section, we demonstrate its capability across four settings.
\subsubsection{Extension from Identical Center Tokens}

\begin{figure}[h]
    \centering
    \includegraphics[width=1\columnwidth]{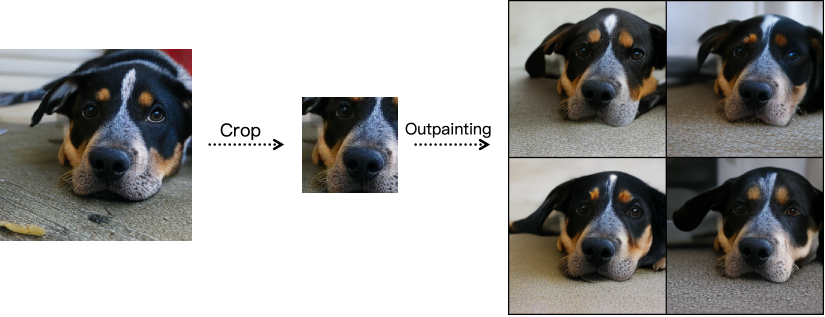}
    \caption{Image extension results based on the same center 8×8 tokens.}
    \label{image_extension2}
\end{figure}

We select a 256×256 image and tokenize it into a 16×16 grid using a VQ-VAE encoder. The central 8×8 tokens, obtained via our spiral token unfolding strategy, are used as the condition input to our EAR model. The model then autoregressively predicts the full 16×16 token map, which is decoded back into the image domain via VQ-VAE. As shown in Figure~\ref{image_extension2}, the generated images effectively reconstruct diverse contents based on the same center region.

\subsubsection{Extension from Different Crops}

\begin{figure}[h]
\centering
\includegraphics[width=0.95\columnwidth]{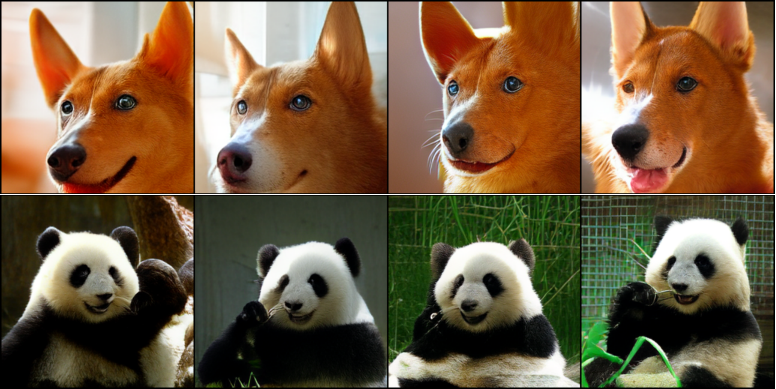} 
\caption{Results of image extension task. The generation depends on the given images ($128\times128$) of the central region.}
\label{image_extension1}
\end{figure}
We apply different cropping and flipping operations on the same ImageNet image to produce multiple $256\times256$ variants. Following the same extension procedure as above, our model generates diverse completions reflecting variations in the input crops, as illustrated in Figure~\ref{image_extension1}.

\subsubsection{Extension from Scaled-down Full Images}

\begin{figure}[h]
    \centering
    \includegraphics[width=1\columnwidth]{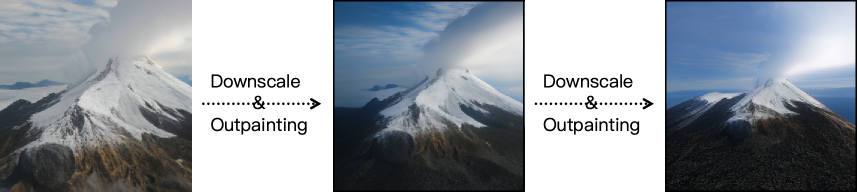}
    \caption{Image extension from downscaled inputs.}
    \label{image_extension3}
\end{figure}

Given an original 256×256 image, we first downscale it to lower resolutions, specifically 192×192 and 128×128. These resized images are then tokenized by the VQ-VAE encoder into 12×12 and 8×8 token grids, respectively. 
We treat these token grids as partial observations and use them as conditional inputs to our EAR model, which autoregressively predicts the full 16×16 token sequence. Finally, the generated token map is decoded back into the image domain using the VQ-VAE decoder, resulting in a complete 256×256 image, as illustrated in Figure~\ref{image_extension3}.

\subsubsection{Cross-Class Image Extension}

\begin{figure}[h]
    \centering
    \includegraphics[width=1\columnwidth]{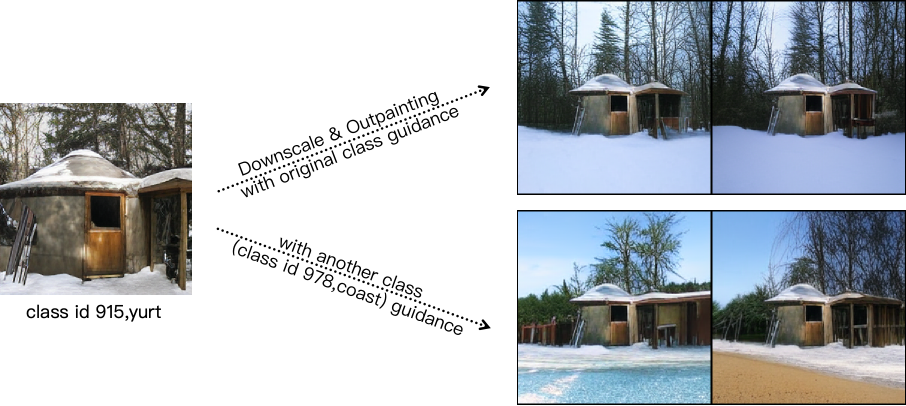}
    \caption{Image extension from downscaled inputs.}
    \label{image_extension4}
\end{figure}


We conducted an interesting cross-class image extension experiment. A 
\(256\times256\) image was downscaled and tokenized with VQ-VAE, and token generation was guided by an embedding from another class. As shown in Figure~\ref{image_extension4}, the original image depicts a yurt in a snowy environment. With the original class embedding, the generated result preserves the snow scene; with the “coast” embedding, the yurt unexpectedly appears on a beach, revealing an intriguing cross-semantic transformation.

\section{Conclusion}

In this paper, we address a key challenge in visual autoregressive modeling: achieving high-fidelity next-token prediction while remaining efficient for high-resolution image synthesis. Inspired by human visual cognition, we propose Expanding Autoregressive Modeling (EAR), which integrates a spiral token unfolding strategy with next-any-token prediction. This center-outward process preserves spatial continuity and enables controlled multi-token parallelization. On ImageNet $256\times256$, EAR surpasses prior autoregressive baselines, generating images substantially faster than LlamaGen~\cite{sun2024autoregressive} and matching or exceeding VAR~\cite{tian2024visualautoregressivemodelingscalable} in quality at lower computational cost. These results show that EAR offers an efficient, cognitively inspired alternative to conventional multi-scale generation.

{
    \small
    \bibliographystyle{ieeenat_fullname}
    \bibliography{main}
}
\clearpage
\appendix  

\twocolumn[  
\begin{center}
    {\Large \textbf{Appendix}}  
    \vspace{4mm}
\end{center}
]

\section{Overall Quantitative Comparison}

\begin{figure}[t]
    \centering
    \includegraphics[width=0.85\linewidth]{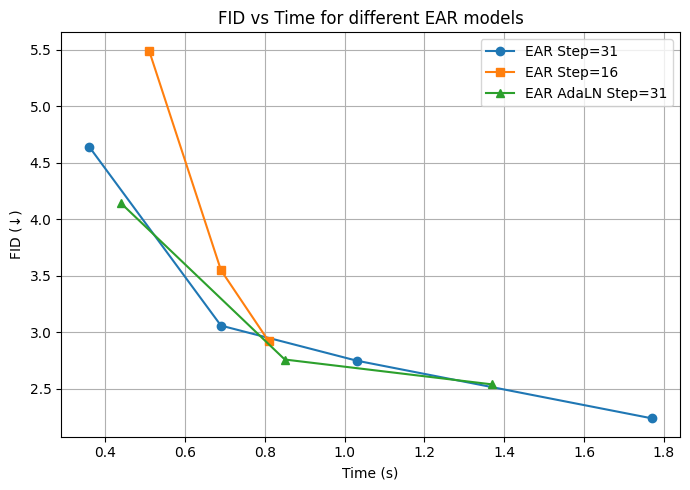}
    \caption{FID comparison with respect to inference time across different EAR model variants. Models with more parameters generally achieve lower FID but require longer inference time. The AdaLN-enhanced models consistently outperform their vanilla counterparts at similar time costs. }
    \label{fid_t}
\end{figure}

\begin{figure}[t]
    \centering
    \includegraphics[width=0.85\linewidth]{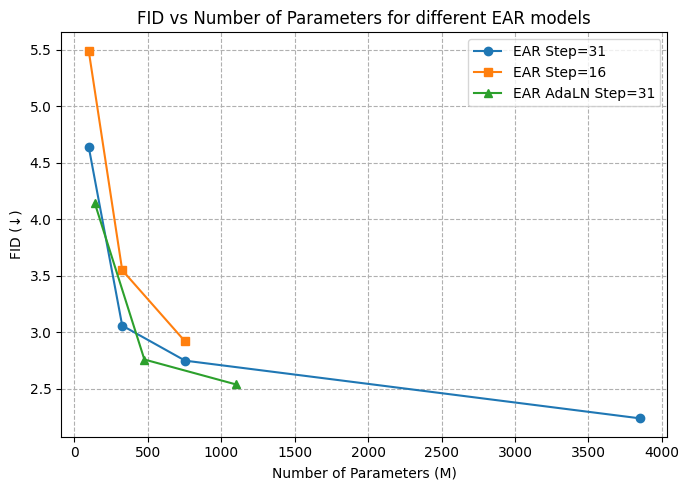}
    \caption{FID comparison with respect to model parameter size. Larger EAR models achieve better generative quality with steadily decreasing FID. The AdaLN variants show improved efficiency by obtaining lower FID with fewer parameters.}
    \label{fid_p}
\end{figure}

Table~\ref{Overall-comparison} and Figure~\ref{fid_t}~\ref{fid_p} present a quantitative comparison of our proposed EAR models and their AdaLN variants on ImageNet 256$\times$256. The metrics include \textbf{FID} (Fréchet Inception Distance, lower is better) and \textbf{IS} (Inception Score, higher is better), along with the number of parameters (\#Para), generation steps (\#Step), and time for 32-image batch (in seconds).

From the results, several observations can be made:

\begin{itemize}
    \item \textbf{Scaling Effect:} Increasing model size from EAR-B to EAR-XXXL consistently reduces FID and increases IS, indicating better image quality and more realistic samples.
    
    \item \textbf{Sampling Steps:} Reducing the number of generation steps (from 31 to 16) slightly increases FID but decreases inference time, demonstrating a trade-off between quality and speed.
    
    \item \textbf{AdaLN Variants:} Incorporating AdaLN generally improves IS while maintaining or slightly reducing FID compared to the baseline models with the same architecture, showing the effectiveness of adaptive normalization in enhancing class-conditional generation.
    
    \item \textbf{Computational Efficiency:} Even for the largest model (EAR-XL(adaLN)), the per-image generation time remains reasonable (1.37 s) with a single generation step, illustrating that our approach scales efficiently.
\end{itemize}

In summary, the table demonstrates that EAR models can generate high-quality images with controllable speed-accuracy trade-offs, and adaptive normalization further enhances class-conditional generation performance.

\begin{table}[t]
\centering
\footnotesize
\begin{tabular}{l c c c c c}
\toprule
\textbf{Model} & \textbf{FID} $\downarrow$ & \textbf{IS} $\uparrow$ & \#Para & \#Step & Time\\
\midrule
EAR-B & 4.64 & 218.7 & 98M & 31 & 0.36 \\
EAR-L & 3.06 & 261.4 & 326M & 31 & 0.69 \\
EAR-XL & 2.75 & 275.1 & 754M & 31 & 1.03 \\
EAR-XXXL & 2.24 & 271.1 & 3.85B & 31 & 1.77 \\
\midrule
EAR-B & 5.49 & 225.2 & 98M & 16 & 0.51\\
EAR-L & 3.55 & 256.44 & 326M & 16 & 0.69\\
EAR-XL & 2.92 & 266.03 & 754M & 16 & 0.81\\
\midrule
EAR-B (adaLN) & 4.14 & 237.7 & 140M & 31 & 0.44 \\
EAR-L (adaLN) & 2.76 & 266.4 & 477M & 31 & 0.85 \\
EAR-XL (adaLN) & 2.54 & 262.7 & 1.1B & 31 & 1.37 \\
\bottomrule
\end{tabular}
\caption{Quantitative comparison of EAR models and their AdaLN variants on ImageNet 256$\times$256. Lower FID and higher IS indicate better generative performance.}
\label{Overall-comparison}
\end{table}

\section{Details of Mask Tokens}

To enable parallel generation, we adopt two types of mask tokens in our framework. Although both serve similar purposes, their implementations differ in practice. We provide detailed explanations below.

\subsection{Class Mask Token}

The class token is directly obtained from a class embedding and naturally serves as a semantic indicator for the target category. It inherently functions as the start token for generation, making it unnecessary to introduce a separate start token. Therefore, the causal masking strategy in the Transformer remains unchanged, as described in the main text.

\subsection{Unified Mask Token}

In contrast, the unified mask token is an additional learnable embedding that does not contain class information. To ensure consistent performance, we slightly modify the model structure (as shown in the main figure of the paper): the class token derived from the class embedding is prepended to both the ground-truth token sequence and the mask token sequence. This ensures that the mask token always has access to class-level information, allowing it to guide class-conditional generation.

Importantly, this modification does not alter the overall model design. The task can still be treated as image generation with $(n \times n + 1)$ tokens, and the additional token has negligible impact on computational efficiency.

\section{Next-any-tokens Prediction Method}

In this section, we explain the training and inference procedures of the proposed \emph{Next-any-tokens Prediction} framework, which is designed for efficient autoregressive latent image generation. Our method generalizes traditional next-token prediction by allowing multiple positions (or ``any tokens'') to be predicted in parallel at each step.

\subsection{Training Procedure}

\begin{algorithm}[ht]
\caption{Next-any-tokens Prediction Train}
\begin{algorithmic}[1]
\Input{latent image $gt$, class token $sos$}
\HyperParam{steps $K$, mask token $mt$, mask $M$}
\State $f = ARmodel(*),gt = \text{spiral}(gt)$
\State $MT = mt.\text{expand}(\text{len}(gt))$
\State $L_0 = \text{concat}(sos, gt, sos, MT)$
\For{$k = 1, \cdots, K$}
    \State $L_k = f(L_{k-1}) \gets M$
\EndFor
\State $logits \gets$ extract $MT$ positions from $L_K$
\State $loss = \text{CrossEntropy}(logits, gt)$\\
\Return{$loss$}
\end{algorithmic}
\end{algorithm}

Algorithm 3 shows the training procedure. The input to the model consists of a latent image $gt$ and a class token $sos$. The latent image is first reordered using a spiral scan (\texttt{spiral(gt)}), which ensures that the autoregressive model generates tokens from the center outwards, mimicking human visual perception.  

A mask token $mt$ is expanded to match the length of the latent sequence, forming a masked sequence $L_0$ concatenated with $sos$, the ground truth latent tokens $gt$, another $sos$, and $MT$ (the mask token sequence). This sequence allows the model to learn to predict masked positions while conditioning on known tokens and class information.

The model $f$ (an autoregressive transformer) is applied iteratively for $K$ steps. At each step, the sequence $L_{k-1}$ is updated according to the mask $M$, gradually refining the prediction of masked positions. After $K$ iterations, the logits corresponding to masked token positions are extracted and used to compute a cross-entropy loss against the ground truth latent image. This loss is then used to optimize the model parameters.

Formally:
\[
L_k = f(L_{k-1}) \gets M, \quad k = 1,\dots,K
\]
\[
\text{loss} = \text{CrossEntropy}(\text{logits at MT positions}, gt)
\]

\subsection{Inference Procedure}

\begin{algorithm}[ht]
\caption{Next-any-tokens Prediction Inference}
\begin{algorithmic}[1]
\Input{class token $sos$}
\HyperParam{steps $K$, tokens per step $(l_k)_{k=1}^K$, mask token $mt$, KV cache $C$}
\State $f = ARmodel(concat(*,*)) \to (*,*)$
\State $L_0 = sos, L=[\ ]$
\For{$k = 1, \cdots, K$}
    \State $MT_k = mt.\text{expand}(l_k)$
    \State $\_, L_k = f(L_{k-1},MT_k)\ and\ (L_{k-1},MT_k)\to C$
    \State $L = \text{queue\_push}(L,L_k)$
\EndFor
\State $L = \text{unspiral}(L)$\\
\Return{generated latent image $L$}
\end{algorithmic}
\end{algorithm}

Algorithm 4 illustrates the inference process. Starting from a class token $sos$, we generate a latent image in $K$ steps. At each step $k$, a mask token sequence $MT_k$ of length $l_k$ is concatenated with the previously generated sequence $L_{k-1}$. The autoregressive model $f$ predicts the next set of tokens, and the output $L_k$ is stored in a queue $L$ for accumulation. A key-value (KV) cache $C$ is maintained to efficiently store transformer attention states across steps, reducing redundant computation.

After completing all $K$ steps, the accumulated sequence $L$ is reordered back to the original spatial layout using \texttt{unspiral(L)}, producing the final latent image.

The inference procedure can be summarized as:
\[
L = \text{queue\_push}(L, f(L_{k-1}, MT_k)), \quad k = 1,\dots,K
\]
\[
\text{output image} = \text{unspiral}(L)
\]



\section{Visualizations of Generation Results}
\begin{figure*}[h]
\centering
\newcommand{\imgwidth}{0.45\textwidth}

\begin{subfigure}{\imgwidth}
    \centering
    \includegraphics[width=\textwidth]{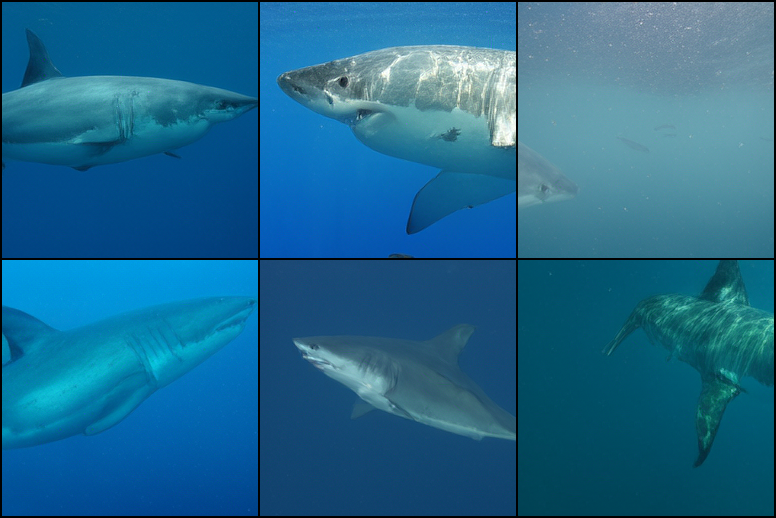}
    \caption*{Class ID: 2}
\end{subfigure}\hfill
\begin{subfigure}{\imgwidth}
    \centering
    \includegraphics[width=\textwidth]{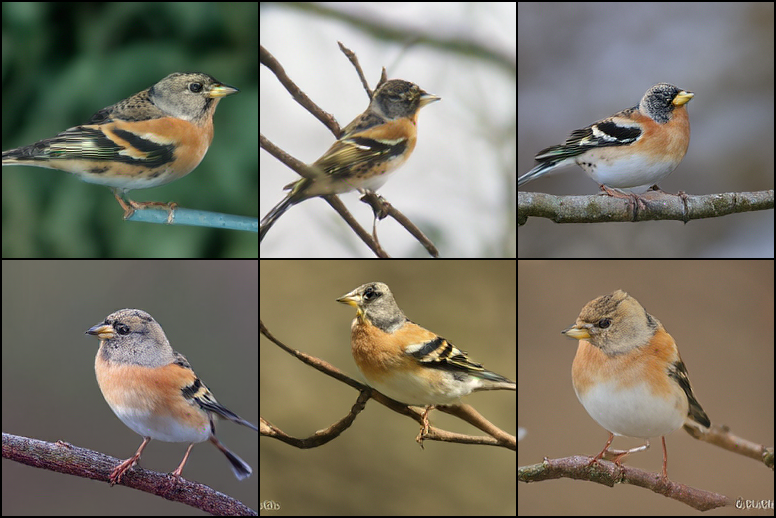}
    \caption*{Class ID: 10}
\end{subfigure}

\vspace{0.6em}

\begin{subfigure}{\imgwidth}
    \centering
    \includegraphics[width=\textwidth]{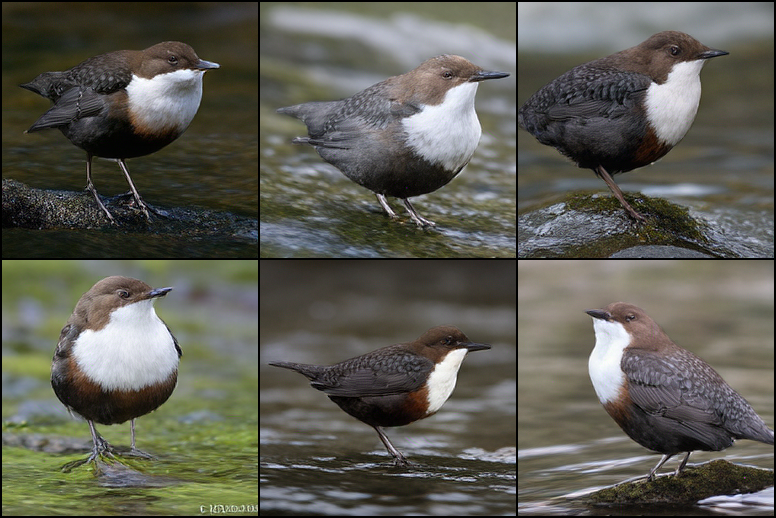}
    \caption*{Class ID: 20}
\end{subfigure}\hfill
\begin{subfigure}{\imgwidth}
    \centering
    \includegraphics[width=\textwidth]{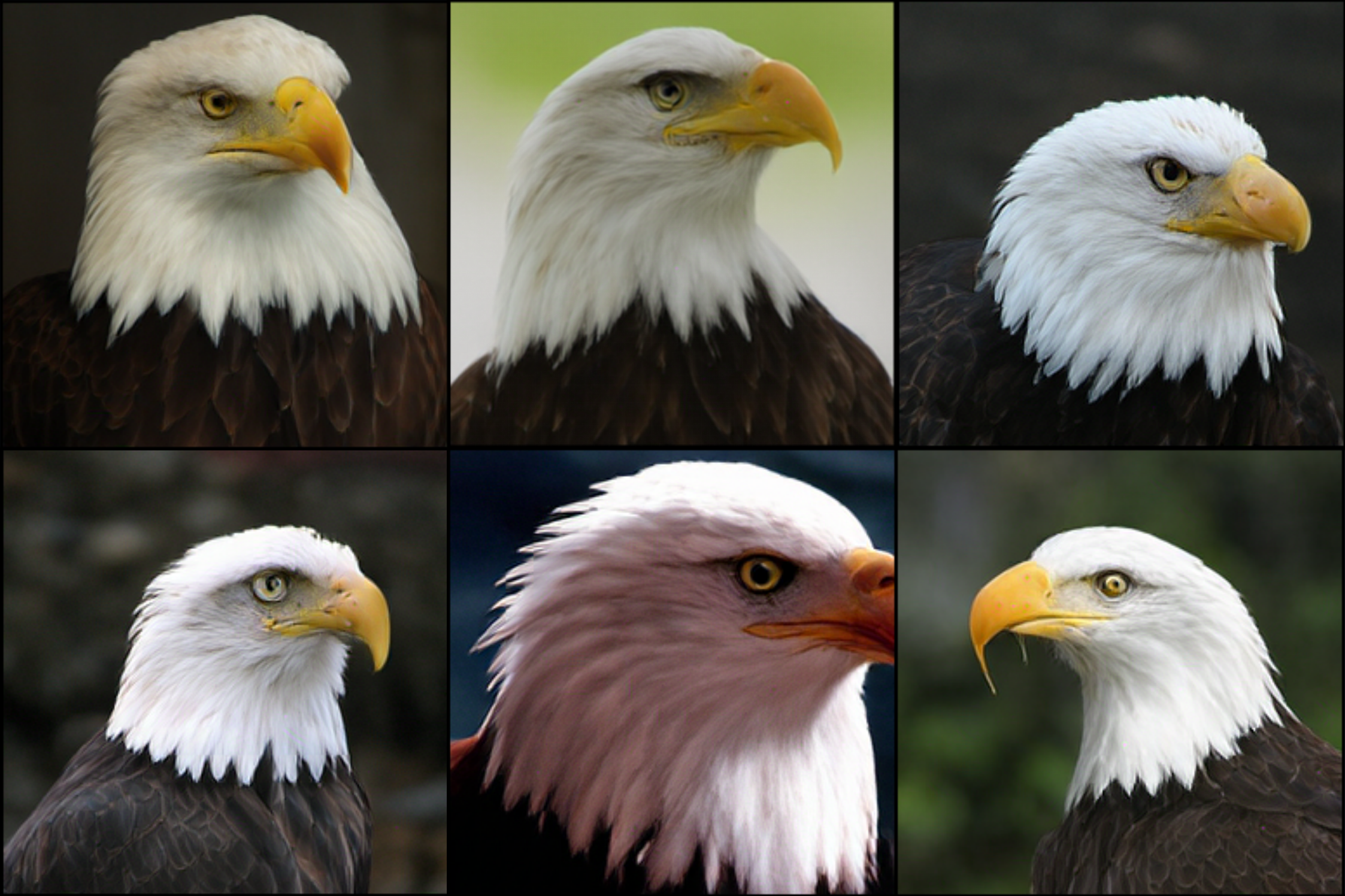}
    \caption*{Class ID: 22}
\end{subfigure}

\vspace{0.6em}

\begin{subfigure}{\imgwidth}
    \centering
    \includegraphics[width=\textwidth]{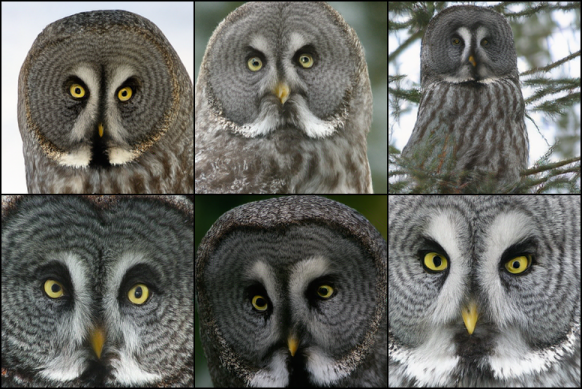}
    \caption*{Class ID: 24}
\end{subfigure}\hfill
\begin{subfigure}{\imgwidth}
    \centering
    \includegraphics[width=\textwidth]{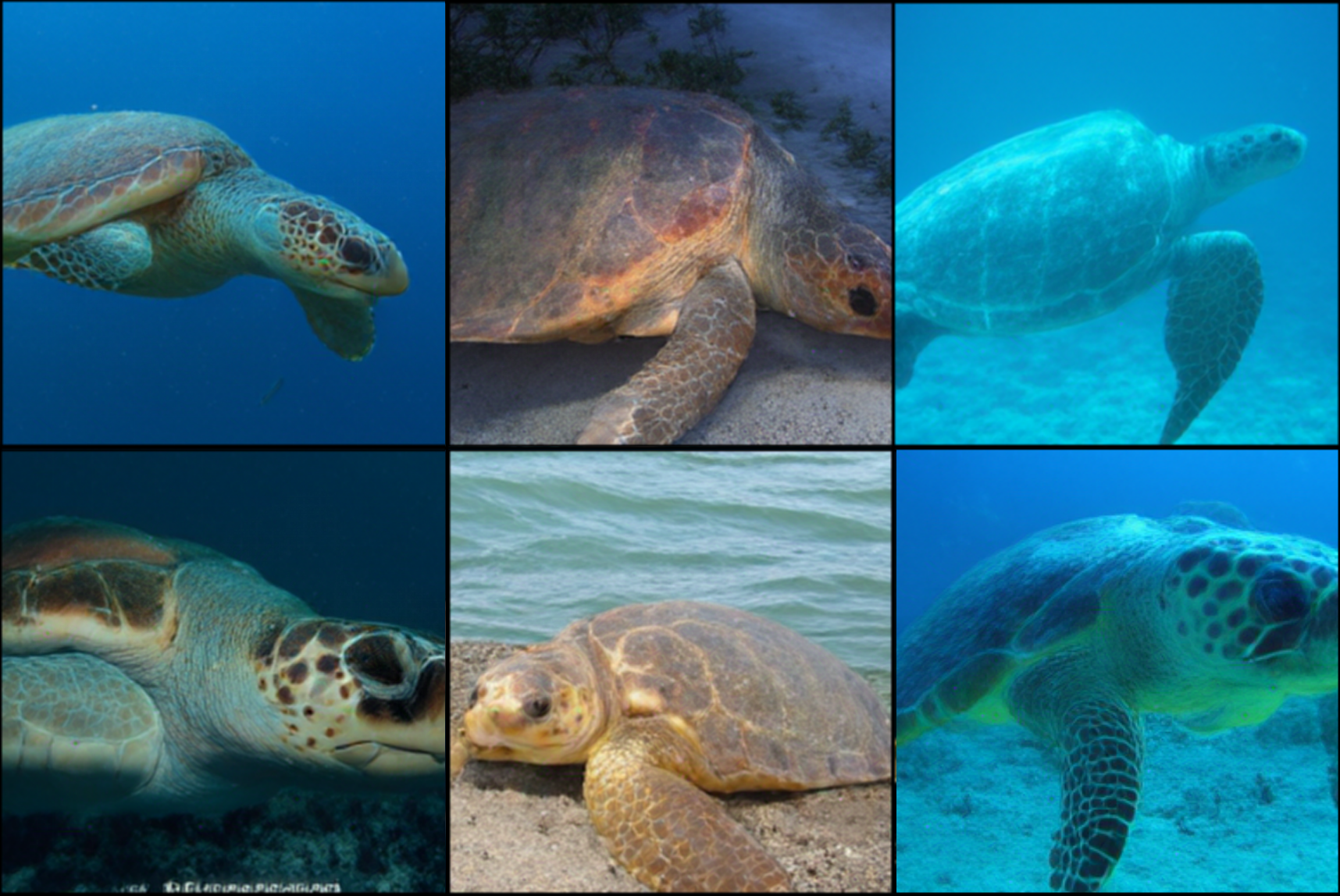}
    \caption*{Class ID: 33}
\end{subfigure}

\caption{Generated images by EAR-XL with AdaLN (Page 1 of 4).}
\label{fig:gen_page1}
\end{figure*}

\begin{figure*}[h]
\centering
\newcommand{\imgwidth}{0.45\textwidth}

\begin{subfigure}{\imgwidth}
    \centering
    \includegraphics[width=\textwidth]{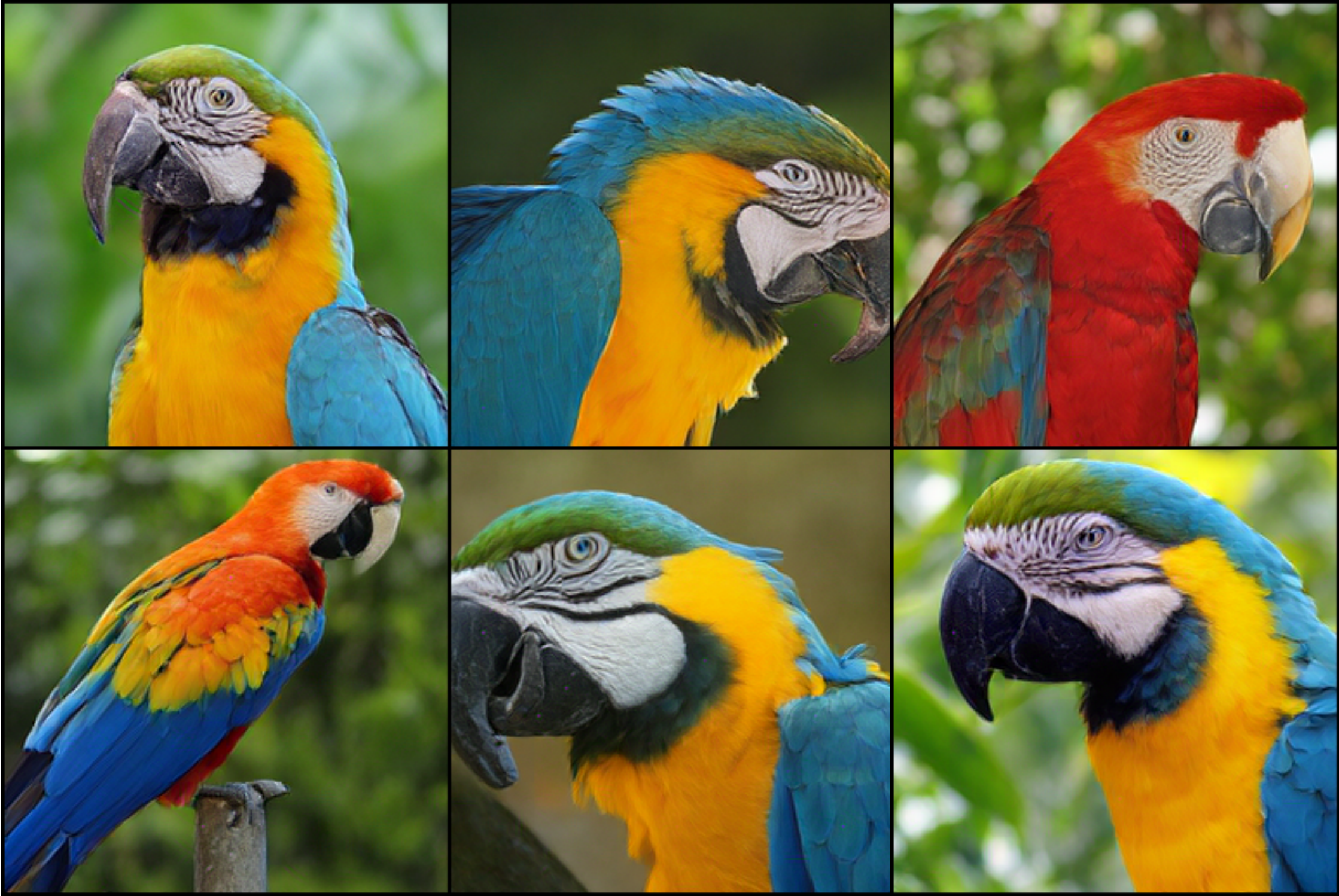}
    \caption*{Class ID: 88}
\end{subfigure}\hfill
\begin{subfigure}{\imgwidth}
    \centering
    \includegraphics[width=\textwidth]{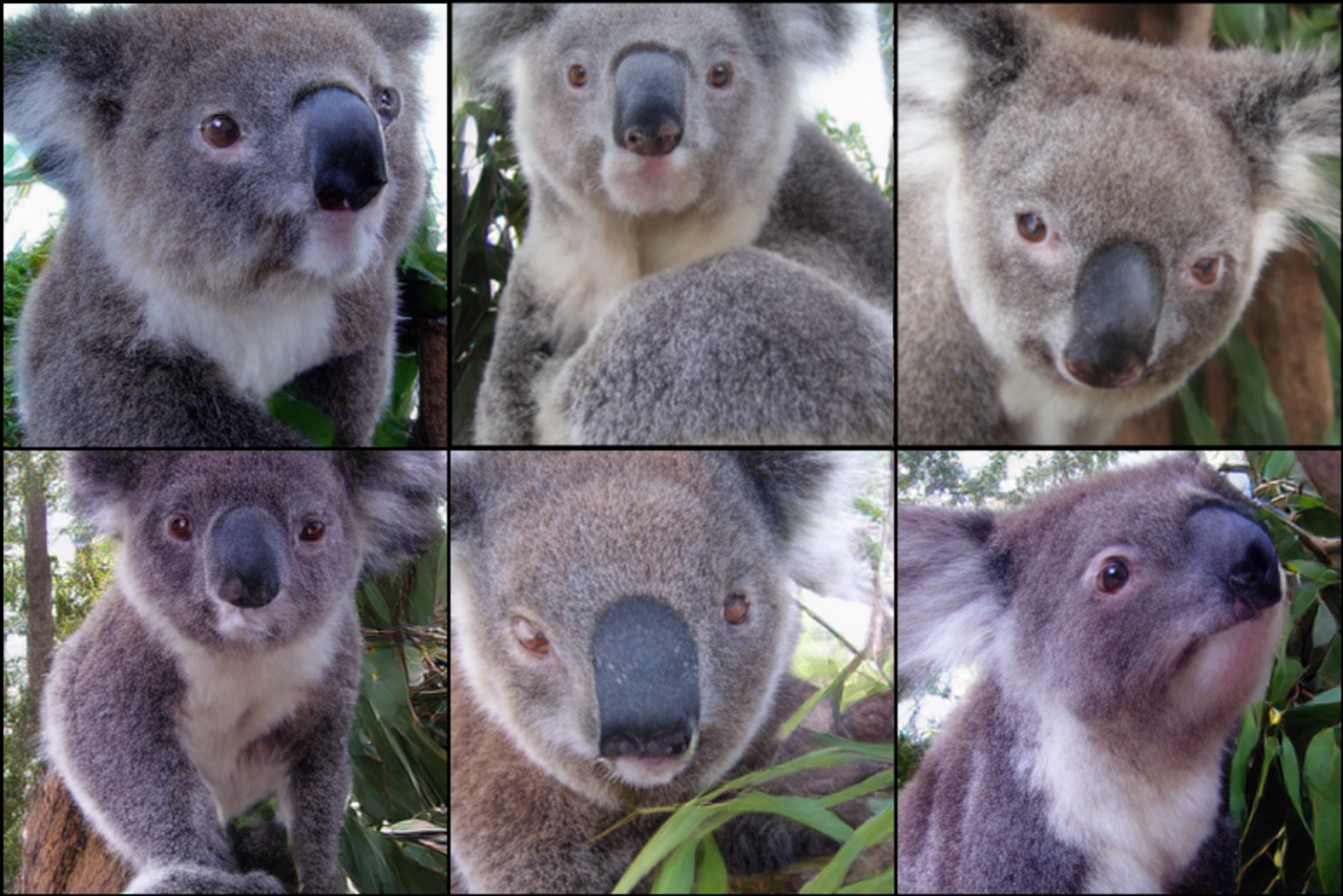}
    \caption*{Class ID: 105}
\end{subfigure}

\vspace{0.6em}

\begin{subfigure}{\imgwidth}
    \centering
    \includegraphics[width=\textwidth]{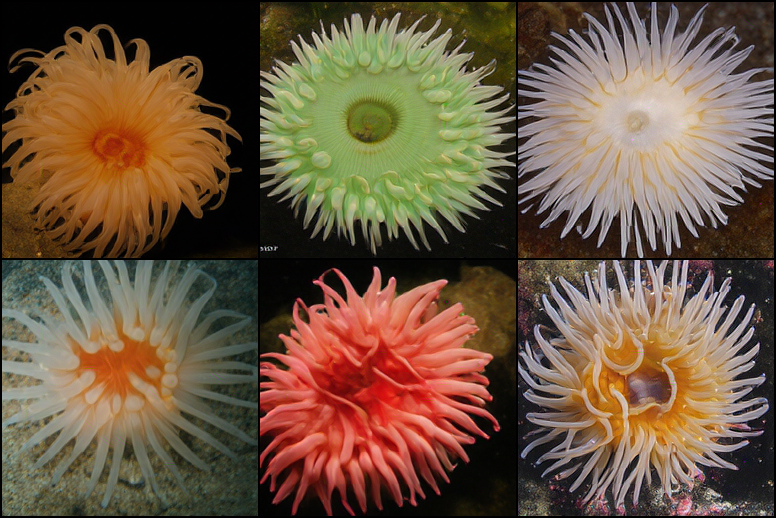}
    \caption*{Class ID: 108}
\end{subfigure}\hfill
\begin{subfigure}{\imgwidth}
    \centering
    \includegraphics[width=\textwidth]{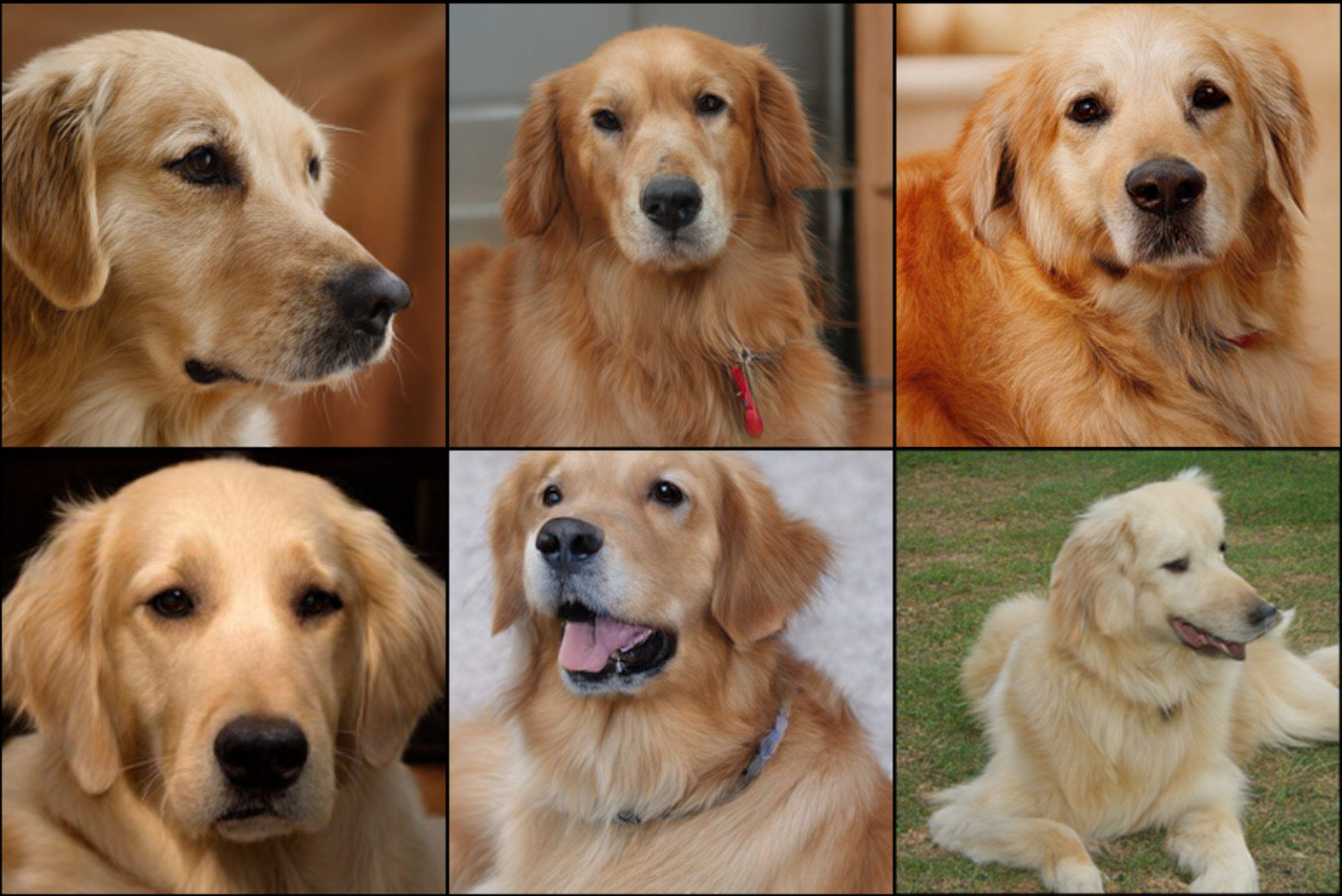}
    \caption*{Class ID: 207}
\end{subfigure}

\vspace{0.6em}

\begin{subfigure}{\imgwidth}
    \centering
    \includegraphics[width=\textwidth]{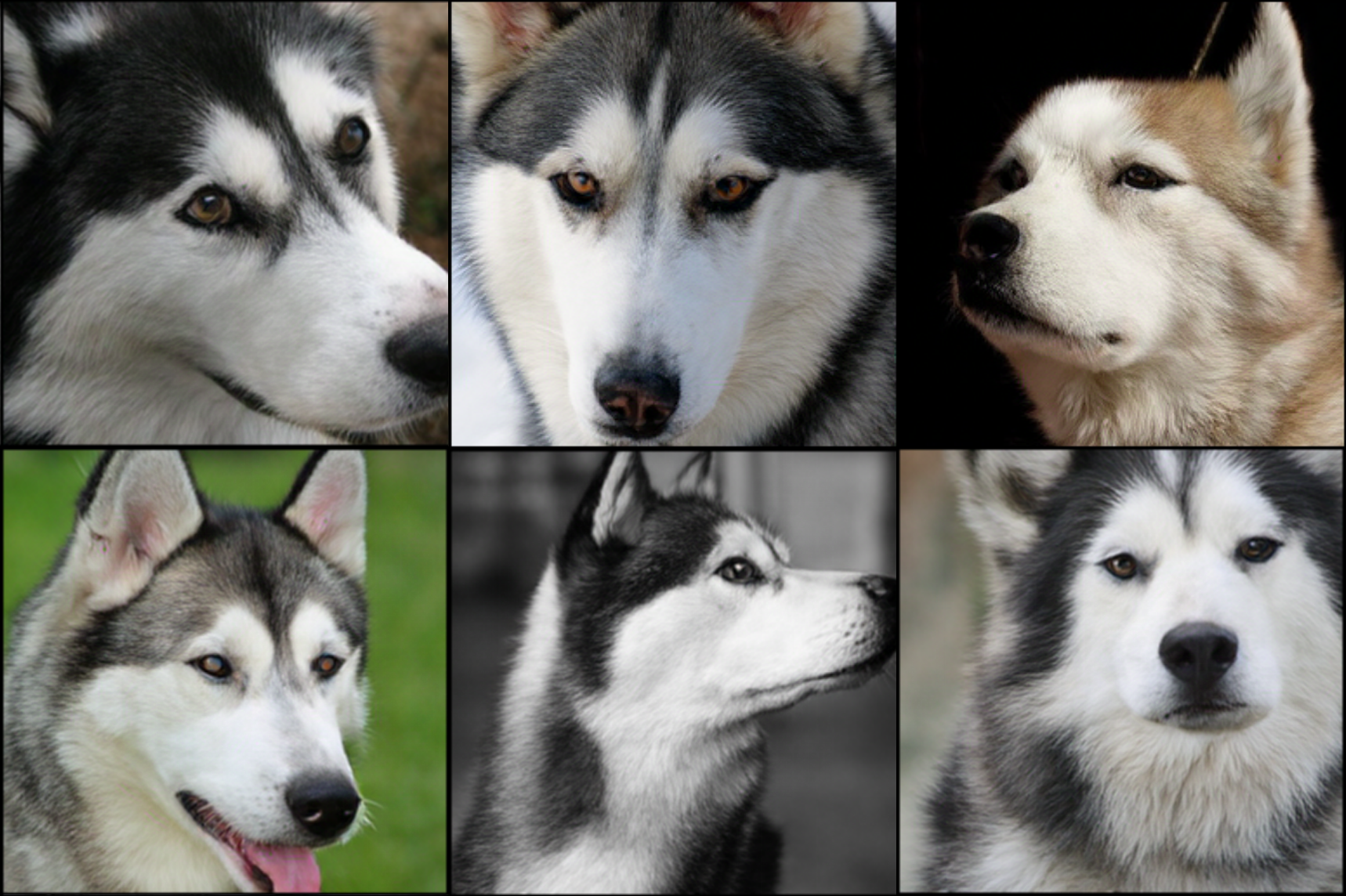}
    \caption*{Class ID: 248}
\end{subfigure}\hfill
\begin{subfigure}{\imgwidth}
    \centering
    \includegraphics[width=\textwidth]{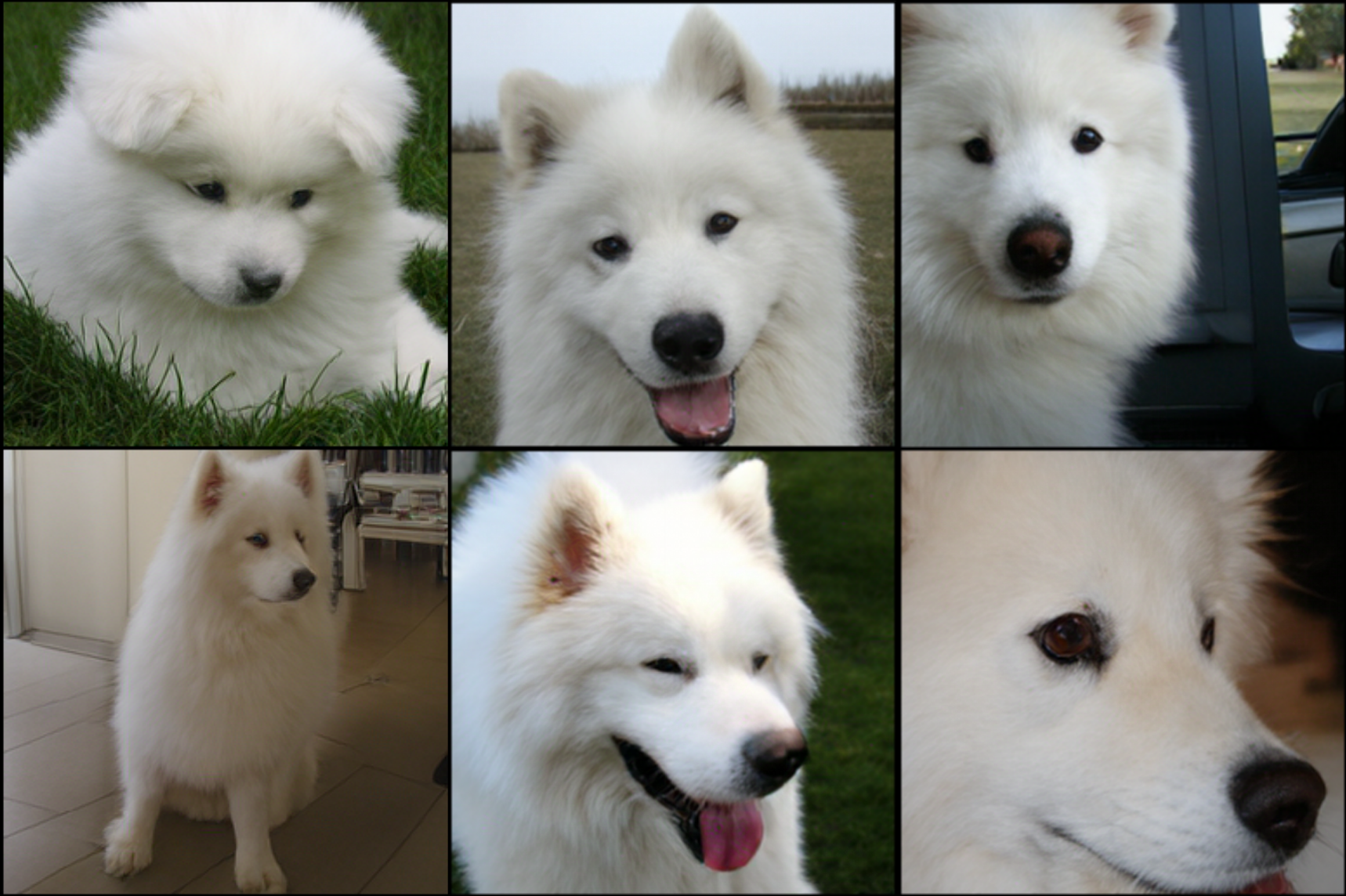}
    \caption*{Class ID: 258}
\end{subfigure}

\caption{Generated images by EAR-XL with AdaLN (Page 2 of 4).}
\label{fig:gen_page2}
\end{figure*}

\begin{figure*}[h]
\centering
\newcommand{\imgwidth}{0.45\textwidth}

\begin{subfigure}{\imgwidth}
    \centering
    \includegraphics[width=\textwidth]{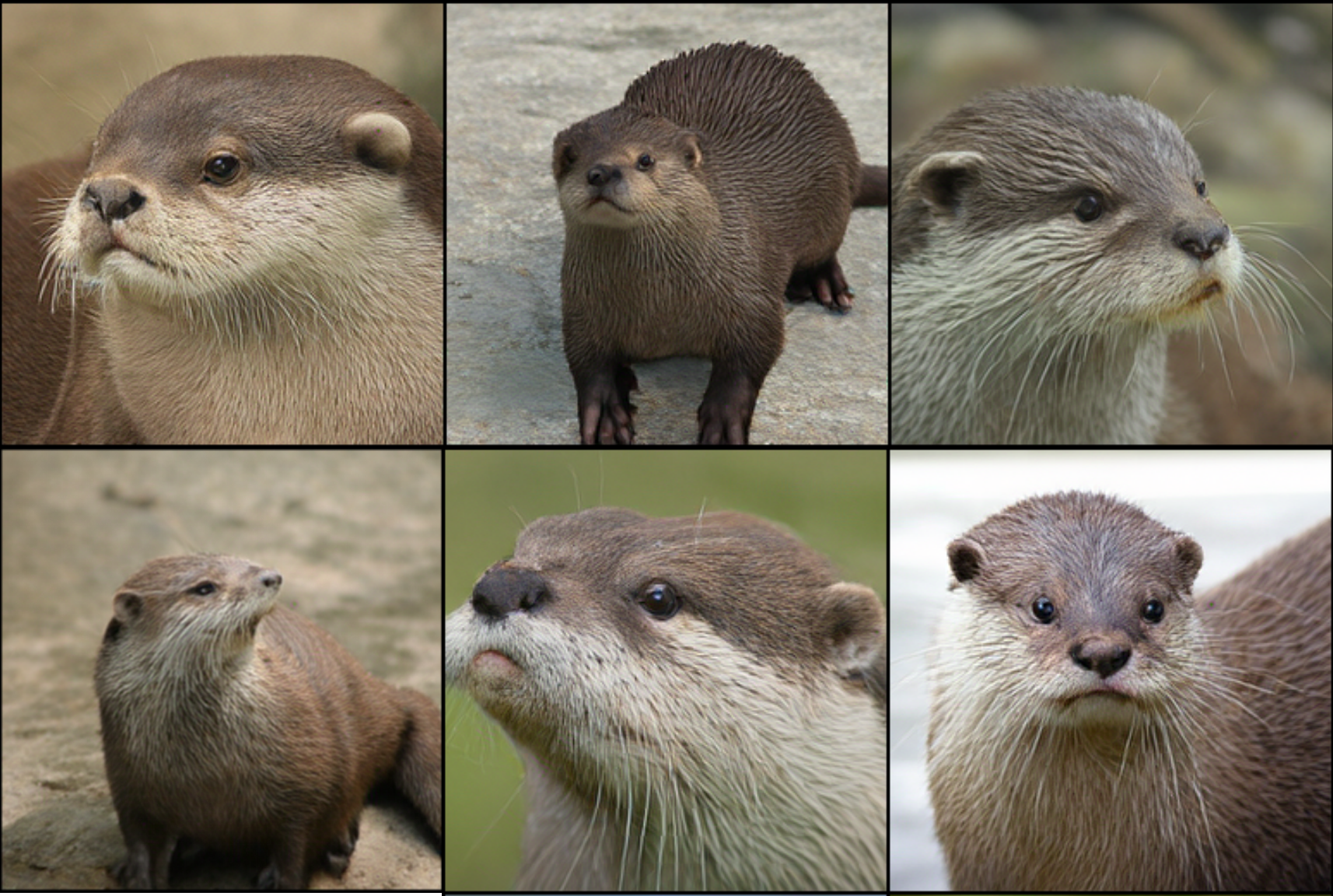}
    \caption*{Class ID: 360}
\end{subfigure}\hfill
\begin{subfigure}{\imgwidth}
    \centering
    \includegraphics[width=\textwidth]{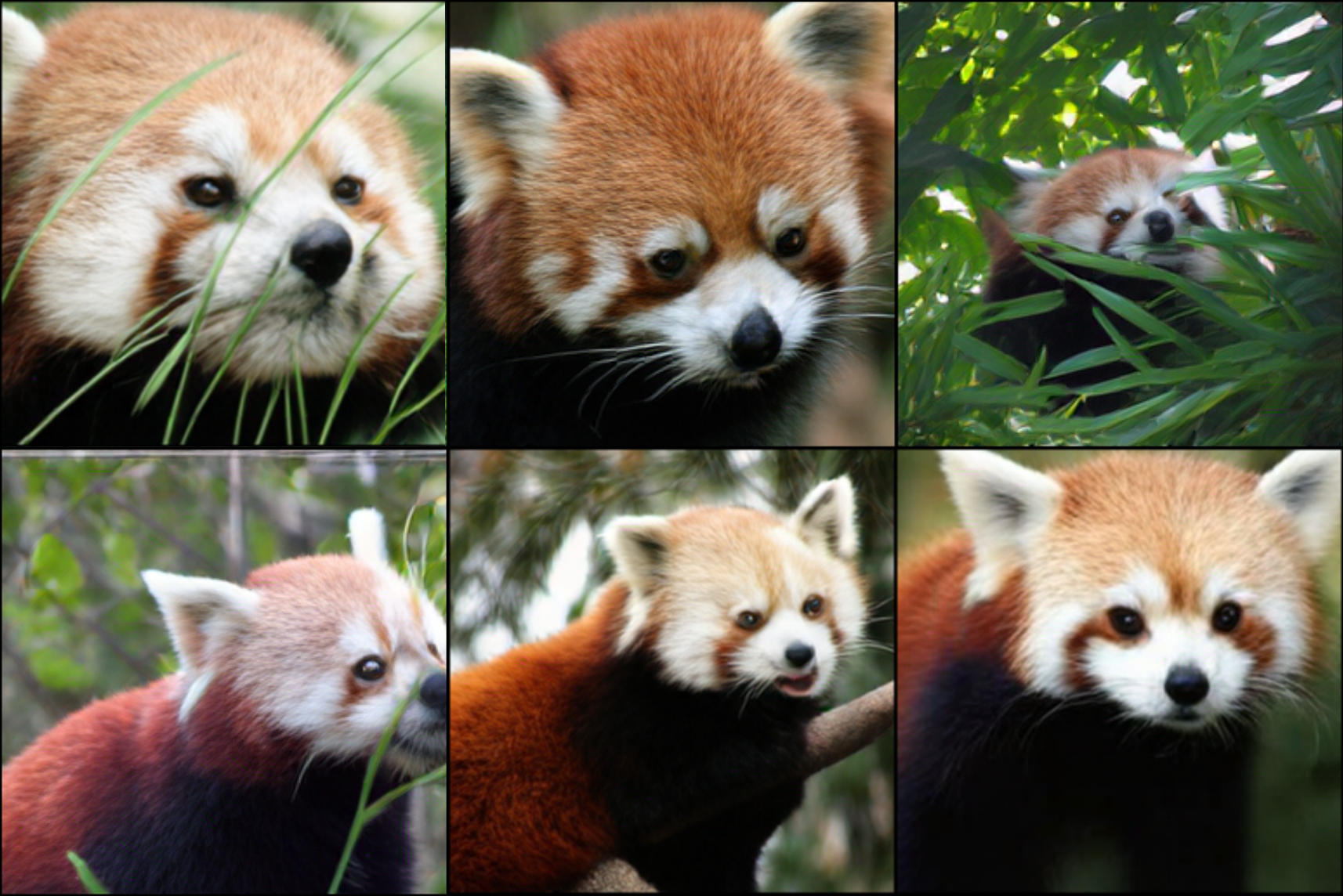}
    \caption*{Class ID: 387}
\end{subfigure}

\vspace{0.6em}

\begin{subfigure}{\imgwidth}
    \centering
    \includegraphics[width=\textwidth]{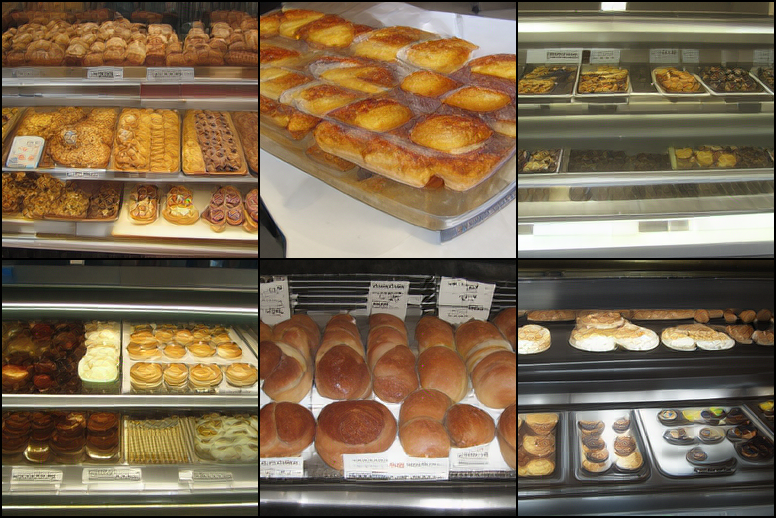}
    \caption*{Class ID: 415}
\end{subfigure}\hfill
\begin{subfigure}{\imgwidth}
    \centering
    \includegraphics[width=\textwidth]{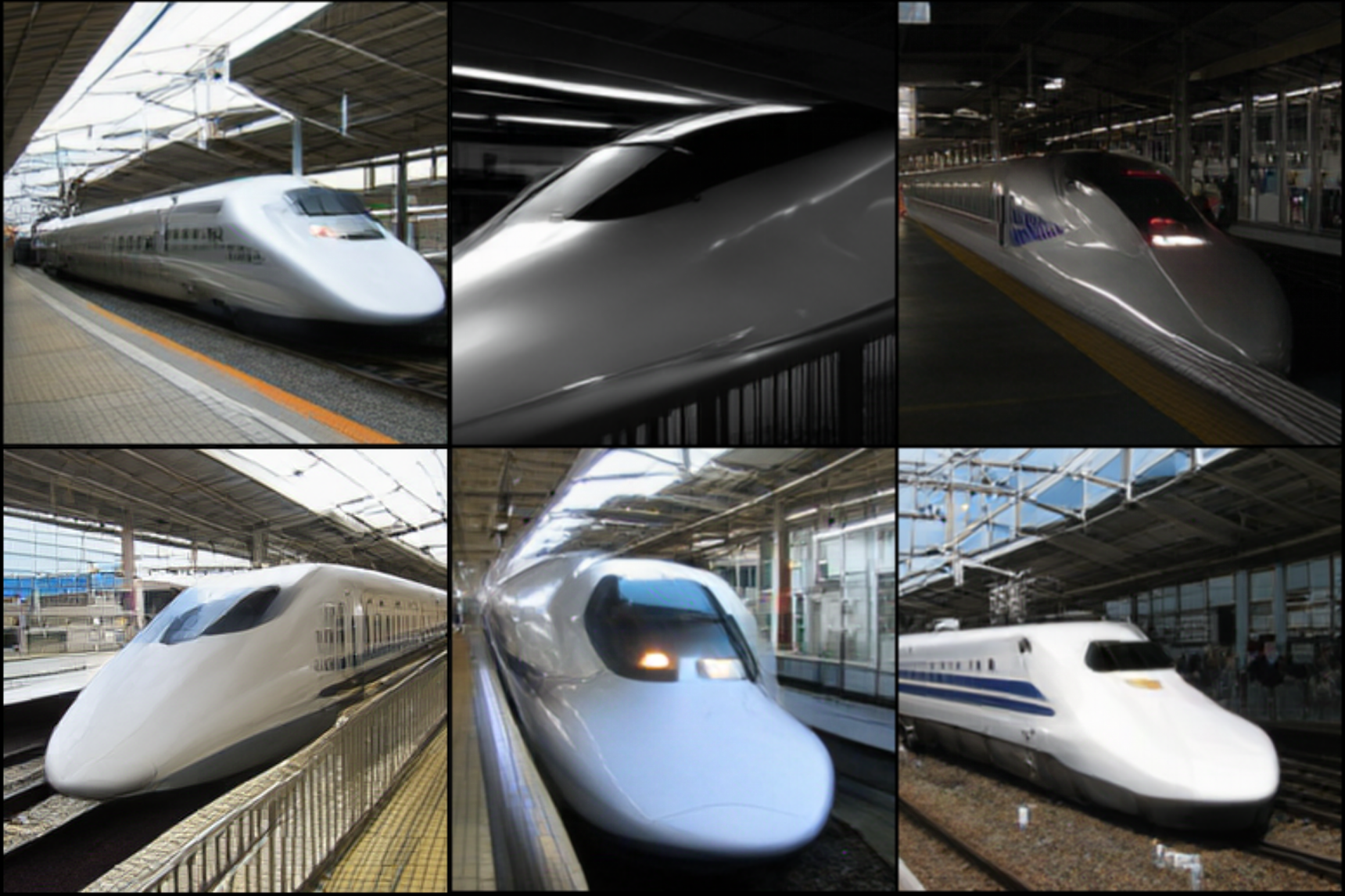}
    \caption*{Class ID: 466}
\end{subfigure}

\vspace{0.6em}

\begin{subfigure}{\imgwidth}
    \centering
    \includegraphics[width=\textwidth]{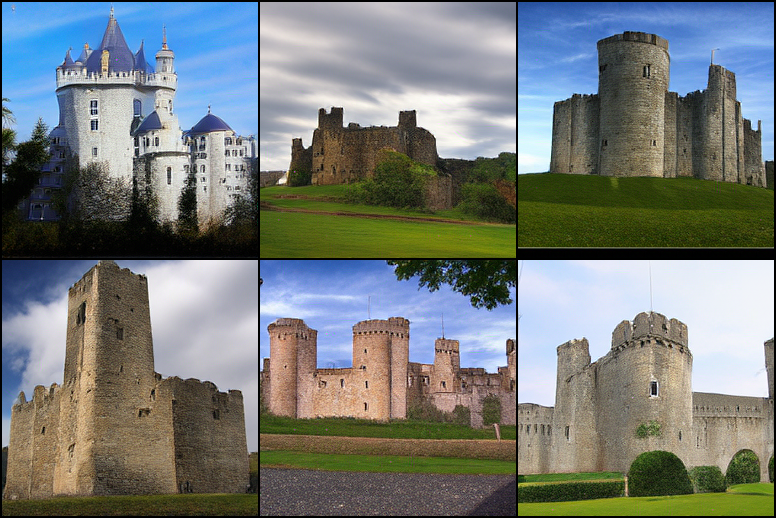}
    \caption*{Class ID: 483}
\end{subfigure}\hfill
\begin{subfigure}{\imgwidth}
    \centering
    \includegraphics[width=\textwidth]{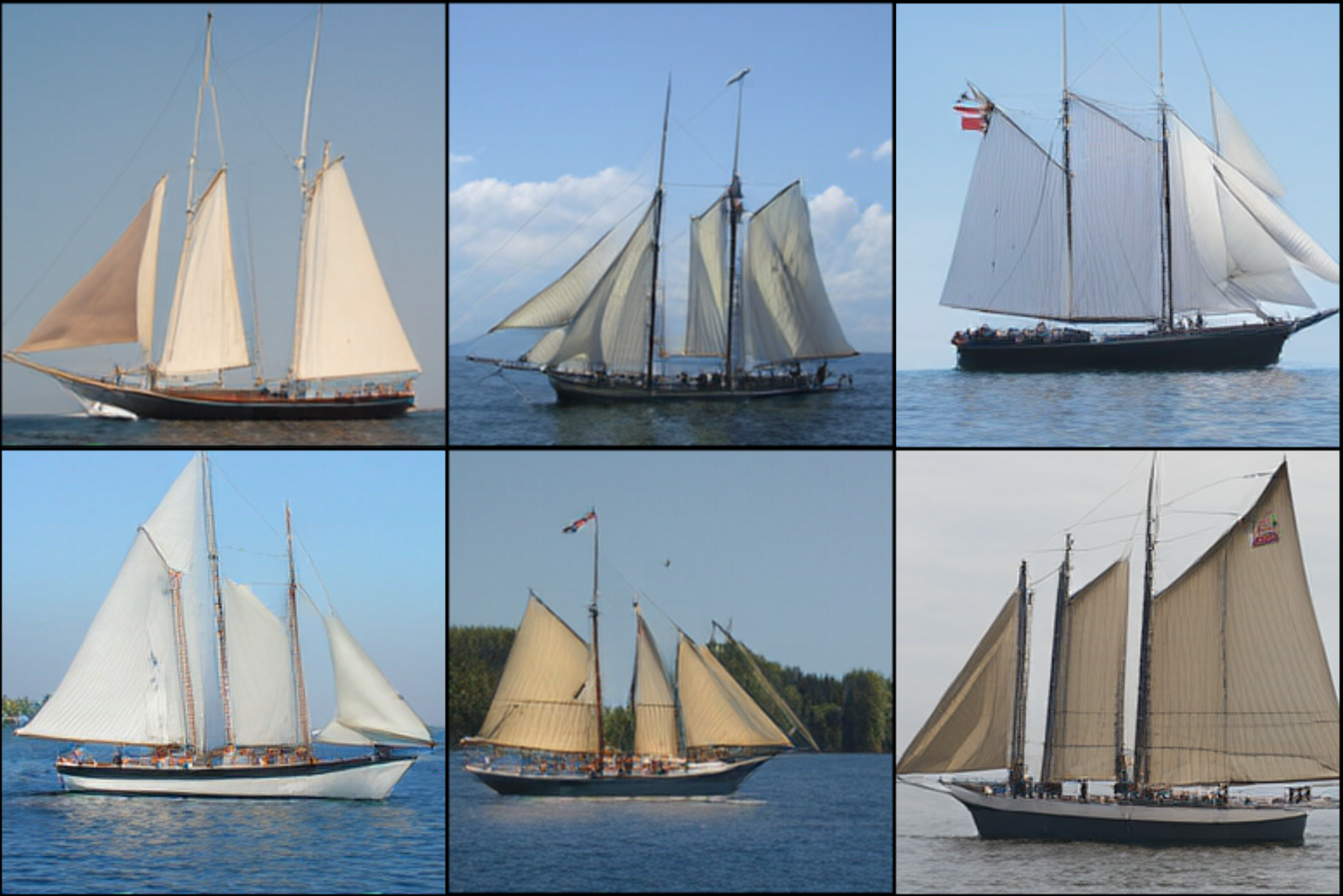}
    \caption*{Class ID: 780}
\end{subfigure}

\caption{Generated images by EAR-XL with AdaLN (Page 3 of 4).}
\label{fig:gen_page3}
\end{figure*}

\begin{figure*}[h]
\centering
\newcommand{\imgwidth}{0.45\textwidth}

\begin{subfigure}{\imgwidth}
    \centering
    \includegraphics[width=\textwidth]{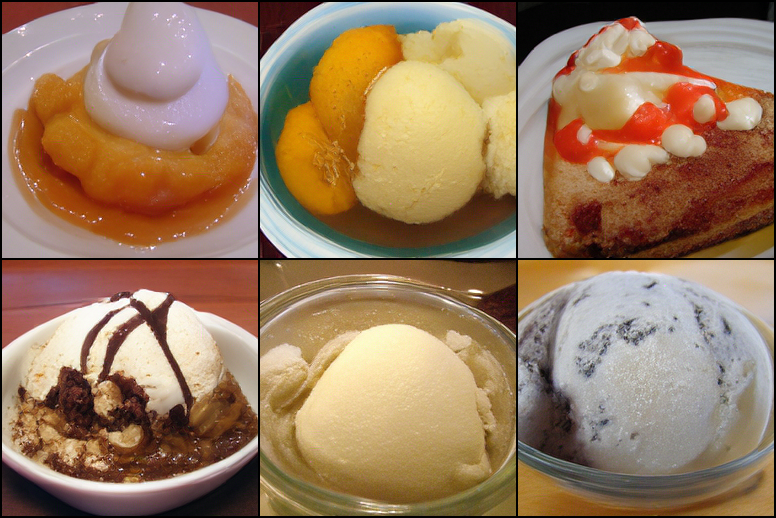}
    \caption*{Class ID: 928}
\end{subfigure}\hfill
\begin{subfigure}{\imgwidth}
    \centering
    \includegraphics[width=\textwidth]{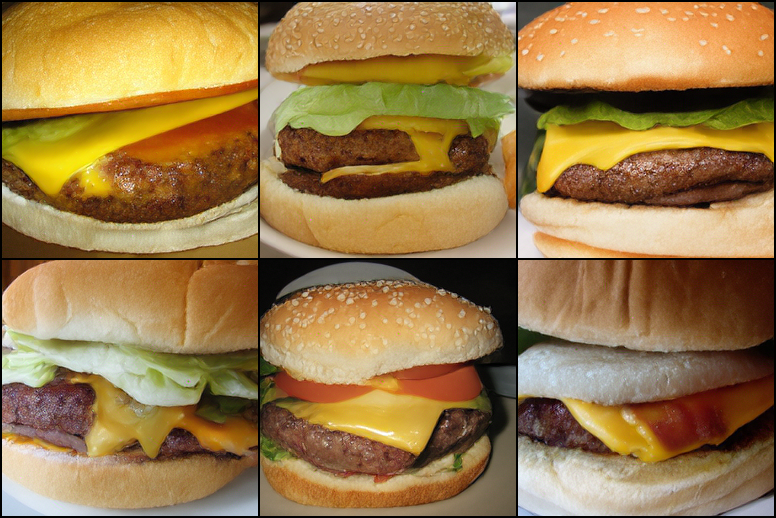}
    \caption*{Class ID: 933}
\end{subfigure}

\vspace{0.6em}

\begin{subfigure}{\imgwidth}
    \centering
    \includegraphics[width=\textwidth]{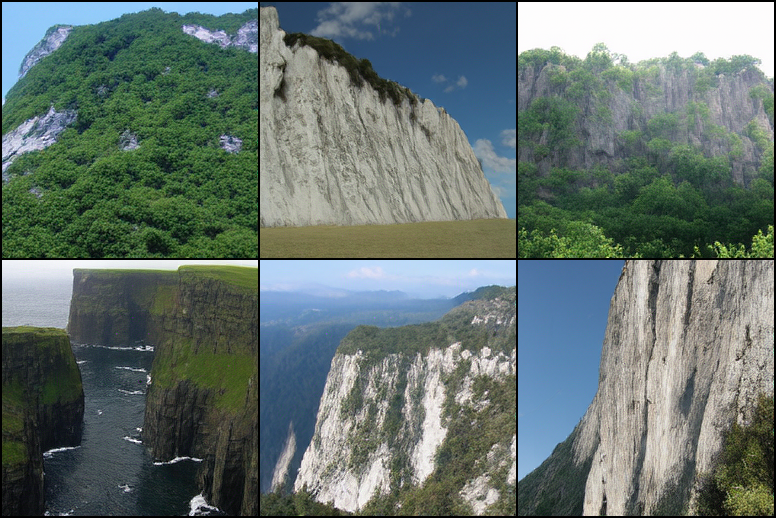}
    \caption*{Class ID: 972}
\end{subfigure}\hfill
\begin{subfigure}{\imgwidth}
    \centering
    \includegraphics[width=\textwidth]{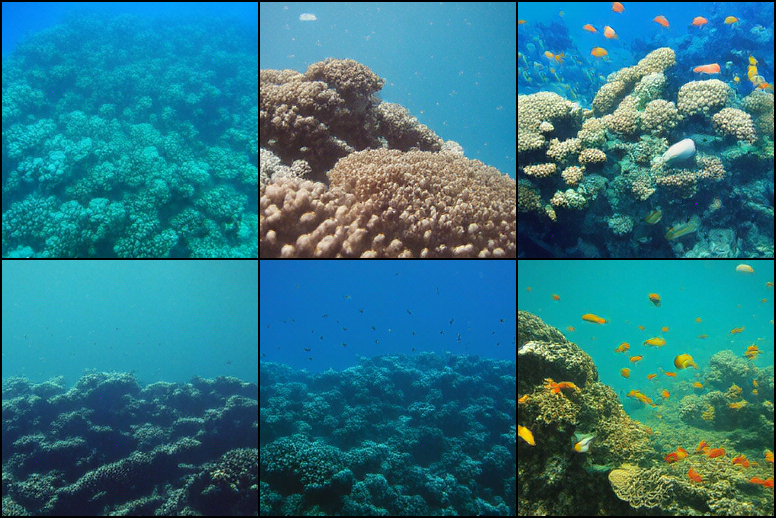}
    \caption*{Class ID: 973}
\end{subfigure}

\vspace{0.6em}

\begin{subfigure}{\imgwidth}
    \centering
    \includegraphics[width=\textwidth]{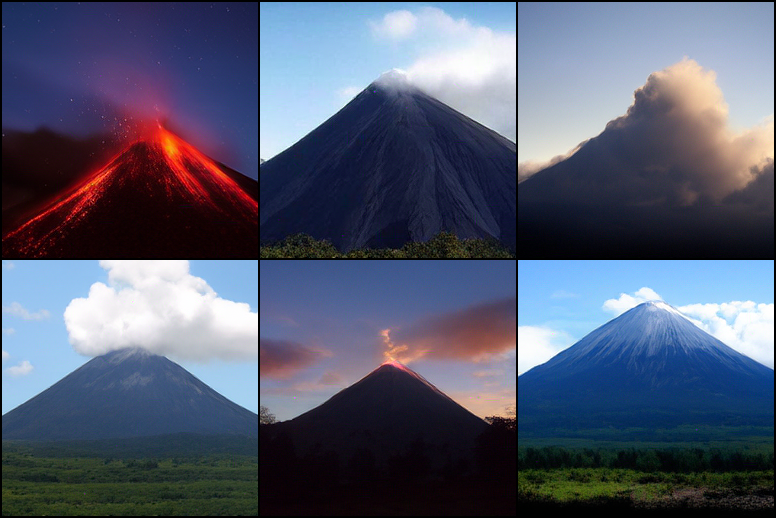}
    \caption*{Class ID: 980}
\end{subfigure}\hfill
\begin{subfigure}{\imgwidth}
    \centering
    \includegraphics[width=\textwidth]{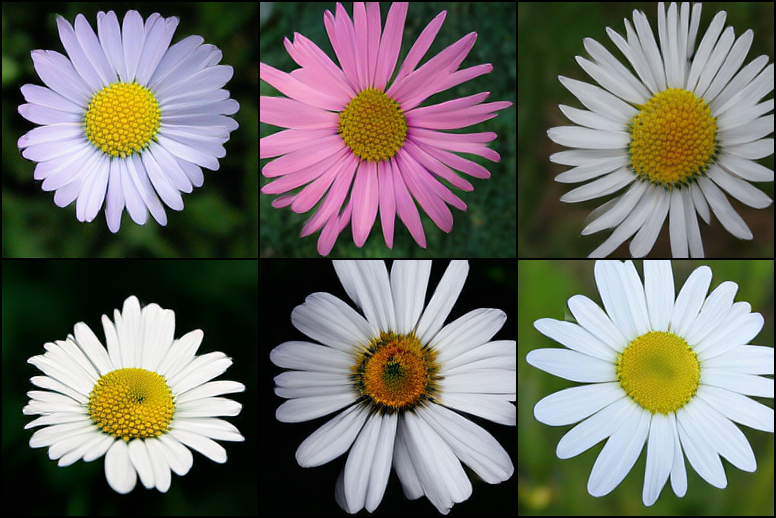}
    \caption*{Class ID: 985}
\end{subfigure}

\caption{Generated images by EAR-XL with AdaLN (Page 4 of 4).}
\label{fig:gen_page4}
\end{figure*}

We visualize the generation results of our EAR on ImageNet $256\times256$ images. As shown in Figure~\ref{fig:gen_page1}--\ref{fig:gen_page4}, the generated images exhibit high visual fidelity with crisp details, coherent object boundaries, and semantically consistent layouts.

\clearpage

\clearpage

\clearpage

\clearpage

\clearpage

\end{document}